\documentclass[preprint,12pt]{elsarticle}

\usepackage{lineno,hyperref}
\usepackage{natbib}
\usepackage{geometry}
\usepackage{fleqn}
\usepackage{graphicx}
\usepackage{newtxtext,newtxmath}
\usepackage{hyperref}
\usepackage{booktabs}
\usepackage{bm}
\usepackage{amsmath}
\usepackage{enumitem}
\usepackage{esvect}
\usepackage{multicol}
\usepackage{multirow}
\usepackage{soul,xcolor}
\modulolinenumbers[5]

\graphicspath{{./figs/}}
\journal{Elsevier}

\begin{document}

\begin{frontmatter}

\title{\large{Functional PCA and Deep Neural Networks-based Bayesian Inverse Uncertainty Quantification with Transient Experimental Data}}

\author[NCSU]{Ziyu Xie}
\author[NCSU]{Mahmoud Yaseen}

\author[NCSU]{Xu Wu\corref{mycorrespondingauthor}}
\cortext[mycorrespondingauthor]{Corresponding author}
\ead{xwu27@ncsu.edu}

\address[NCSU]{Department of Nuclear Engineering, North Carolina State University    \\ 
	Burlington Engineering Laboratories, 2500 Stinson Drive, Raleigh, NC 27695 \\}

\begin{abstract}
Inverse UQ is the process to inversely quantify the model input uncertainties based on experimental data. This work focuses on developing an inverse UQ process for time-dependent responses, using dimensionality reduction by functional principal component analysis (PCA) and deep neural network (DNN)-based surrogate models. The demonstration is based on the inverse UQ of TRACE physical model parameters using the FEBA transient experimental data. The measurement data is time-dependent peak cladding temperature (PCT). Since the quantity-of-interest (QoI) is time-dependent that corresponds to infinite-dimensional responses, PCA is used to reduce the QoI dimension while preserving the transient profile of the PCT, in order to make the inverse UQ process more efficient. However, conventional PCA applied directly to the PCT time series profiles can hardly represent the data precisely due to the sudden temperature drop at the time of quenching. As a result, a functional alignment method is used to separate the phase and amplitude information of the transient PCT profiles before dimensionality reduction. DNNs are then trained using PC scores from functional PCA to build surrogate models of TRACE in order to reduce the computational cost in Markov Chain Monte Carlo sampling. Bayesian neural networks are used to estimate the uncertainties of DNN surrogate model predictions. In this study, we compared four different inverse UQ processes with different dimensionality reduction methods and surrogate models. The proposed approach shows an improvement in reducing the dimension of the TRACE transient simulations, and the forward propagation of inverse UQ results has a better agreement with the experimental data.
\end{abstract}


\end{frontmatter}


\section{Introduction}
\label{section:Introduction}

Uncertainty Quantification (UQ) is the process to quantify the uncertainties in Quantity-of-Interest (QoIs) by propagating the uncertainties in input parameters through a computer model. In the field of nuclear engineering, most UQ research has focused on forward UQ (FUQ), which involves propagating input uncertainties through computational models to quantify uncertainties in QoIs. However, in FUQ, the input parameter uncertainties are often user-defined or based on subjective expert opinion, which lacks mathematical rigor and can introduce inaccuracies. To address this issue, inverse UQ (IUQ) has been developed in order to quantify input uncertainties based on experimental data.  IUQ research in nuclear engineering primarily relies on statistical analysis and the developed methods can be categorized into three groups \cite{wu2021comprehensive}: frequentist (deterministic) \cite{de2012circe} \cite{cacuci2010sensitivity} \cite{shrestha2016inverse} \cite{hu2016inverse} \cite{petruzzi2019casualidad} \cite{damblin2023generalization}, Bayesian (probabilistic) \cite{wu2018inversePart1} \cite{wu2018inversePart2} \cite{wu2018Kriging} \cite{wicaksono2018bayesian} \cite{damblin2020bayesian} \cite{liu2021uncertainty} \cite{robertson2022treating}, and empirical (design-of-experiments) \cite{kovtonyuk2017development} \cite{skorek2017input} \cite{joucla2008dipe} \cite{zhang2019development}. These methods compare computational simulations with experimental data to estimate the uncertainties of the model input parameters. Frequentist IUQ gives the most likely input parameters that can reproduce the experimental data. Bayesian IUQ quantifies the uncertainties of the input parameters by reducing the disagreement between simulation and experimental data. Empirical IUQ seeks a range of input values based on which the model predictions can envelop the measurement data. See \cite{wu2021comprehensive} for a more detailed review and comparison of these approaches.

In addition, there has been a growing interest in IUQ research over the past decade in the nuclear engineering area. For example, multiple international activities have been undertaken to develop and evaluate the effectiveness of IUQ methods. In fact, many of the IUQ methods mentioned above are developed and/or improved within these international projects. Notable among these is the Post-BEMUSE Reflood Models Input Uncertainty Methods (PREMIUM) \cite{reventos2016premium} benchmark, which focuses on core reflood problems and employs Flooding Experiments with Blocked Arrays (FEBA) tests to quantify and validate input uncertainties in system thermal-hydraulics (TH) models. The OECD/NEA has also performed two follow-up projects: the Systematic Approach for Input Uncertainty Quantification Methodology (SAPIUM) \cite{baccou2020sapium} and Application Tests for Realization of Inverse Uncertainty Quantification and Validation Methodologies in thermal-hydraulics (ATRIUM) \cite{ghione2023applying} (ongoing). These projects aim to develop a systematic approach for quantifying and validating the uncertainty of physical models in system TH codes. In this paper, we will focus on the Bayesian IUQ method. Several improvements will be developed and implemented based on our previous work on the modular Bayesian approach \cite{wu2018inversePart1} \cite{wu2018inversePart2}.

In Bayesian IUQ, Markov Chain Monte Carlo (MCMC) methods \cite{andrieu2008tutorial} are usually utilized to explore the posterior distributions of input parameters. MCMC generates samples that follow a probability density proportional to the parameter posterior distribution. However, a typical MCMC algorithm often requires more than 10,000 samples to reach a converged solution. This can be computationally expensive, especially for nuclear Thermal-Hydraulic (TH) system codes. To address this challenge, surrogate models can be employed to significantly reduce the computational cost. Surrogate models give an approximation of the relation of the input and output of the original computer model (also called full model), and they require only a limited number of full model runs for the training process. Some machine learning (ML) methods such as Gaussian process (GP) and deep neural network (DNN) have been widely used as surrogate models. 


The application of surrogate models to replace the original computational models introduces an additional source of uncertainty, referred to as the code or interpolation uncertainty \cite{wu2021comprehensive} \cite{kennedy2001bayesian} in literature. Conventional DNN-based surrogate models give deterministic predictions of the QoIs for given inputs. Consequently, when used as surrogates, DNNs do not provide estimation of the code/interpolation uncertainty directly. To capture the approximation uncertainty introduced by using DNN-based surrogate models, in this study we will implement the Bayesian inference method for UQ of DNNs. Specifically, Bayesian neural networks (BNNs) are trained as surrogate models of the full model. A BNN is a neural network with distributions over parameters. In BNNs, a prior distribution is specified upon the parameters (weights, bias) of neural networks and then, given the training data, the posterior distributions over the parameters are computed, which are used to quantify predictive uncertainty. Our previous work \cite{yaseen2023quantification} benchmarked three methods, namely, Monte Carlo dropout, deep ensembles, and BNN to estimate the prediction/approximation uncertainties of DNNs. In another study \cite{yaseen2023neural}, these methods were applied to time series data derived from TRACE simulations of the FEBA experiments. In this work, the quantified DNN prediction uncertainties, which are essentially the code/interpolation uncertainties when using DNN as surrogate models, will be incorporated into the Bayesian IUQ process.

When performing IUQ for transient problems, the responses typically exhibit time-dependence, resulting in high-dimensional and highly-correlated data. Such high dimensionality and correlation can lead to challenges for surrogate modeling techniques such as GP \cite{wu2018Kriging} and DNN \cite{labbe2009learning}. To overcome this challenge, dimensionality reduction methods such as principal component analysis (PCA) are often employed, and they have shown successful applications in nuclear engineering. In a study by Wu et al. \cite{wu2018Kriging}, the dimensionality of a time-dependent fission gas release model was reduced using PCA. The experimental data was transferred into the Principal Component (PC) subspace within a Bayesian IUQ framework. Similarly, Roma et al. \cite{roma2021bayesian} also utilized PCA in an IUQ study. It's worth noting that PCA has been employed in other areas, such as global sensitivity analysis \cite{wicaksono2016global}.

However, conventional PCA may not accurately represent time series data when the transient profiles contain important phase and magnitude information that need to be preserved simultaneously. The standard PCA technique is applied to centered data, which may smooth out the phase and magnitude information of the dataset. To address this limitation, a functional PCA method has been developed, specifically designed to handle time series data and preserve both phase and amplitude information \cite{tucker2013generative} \cite{tucker2014functional}. By separating the phase and amplitude information of transient data, the functional PCA method overcomes the challenges posed by conventional PCA, enabling more accurate representation and preservation of the essential features in time series data. This functional PCA approach has shown successful applications in the field of TH system code predictions \cite{perret2019global}, demonstrating significant improvements in dimensionality reduction performance. 

This work focus on developing a Bayesian IUQ process for time-dependent QoIs with a demonstration example using the FEBA experimental data and the TRACE computer model. We implemented and compared four different Bayesian IUQ processes: (1) conventional PCA with GP surrogate model as a reference solution, as it is one of the most widely used approach in literature for transient data \cite{wu2018Kriging} \cite{wicaksono2018bayesian}, (2) conventional PCA with DNN surrogate model without code uncertainty, (3) functional PCA with DNN surrogate model without code uncertainty, and (4) functional PCA with DNN surrogate model with code uncertainty through implementation of BNN. Previously, relevant work has been preformed using FEBA experimental data in Bayesian IUQ \cite{perret2022global}, but without considering the surrogate model uncertainty. \textit{The contribution and novelty in this work can be summarized as}: (i) implementation of functional PCA for time-dependent QoI that contains important phase and magnitude information to be preserved during dimensionality reduction, (ii) surrogate modelling with DNN, while accounting for the code/interpolation uncertainties through BNN, solved with variational inference, (3) a comprehensive and systematic investigation of four Bayesian IUQ processes to study the influence of GP/DNN as surrogate models, conventional vs. functional PCAs, as well as the code/interpolation uncertainties introduced by surrogate models.

A complete IUQ study usually includes sensitivity analysis to select the most influential calibrated parameters \cite{trucano2006calibration}, as well as FUQ and validation to test the IUQ results. In this study, we leveraged the sensitivity analysis study in our previous work \cite{xie2022bayesian-physor}. It has been found that functional PCA improves the dimensionality reduction by a better reconstruction quality using only a few PCs. Using the functional PCA and DNN-based surrogate models while accounting for the code/interpolation uncertainty leads to the best Bayesian IUQ results. Forward propagation of the IUQ results and validation using experimental data not seen in IUQ have shown that the proposed approach has the best agreement with experimental data when compared with the other IUQ methods.

The rest of the paper is arranged as follows. Section \ref{section:Problem} gives an introduction to the FEBA experiment and the TRACE computer model. Section \ref{section:Methods} introduces the PCA method with functional alignment, the method used for UQ of DNN model and the Bayesian IUQ methods. Section \ref{section:Results} presents the results for functional PCA, surrogate modeling, various IUQ methods, forward propagation of the IUQ results and validation. Section \ref{section:Conclusions} concludes the paper and discusses the future work.

\section{Problem Definition}
\label{section:Problem}

\begin{figure}[!htb]
	\centering
	\includegraphics[width=0.9\textwidth]{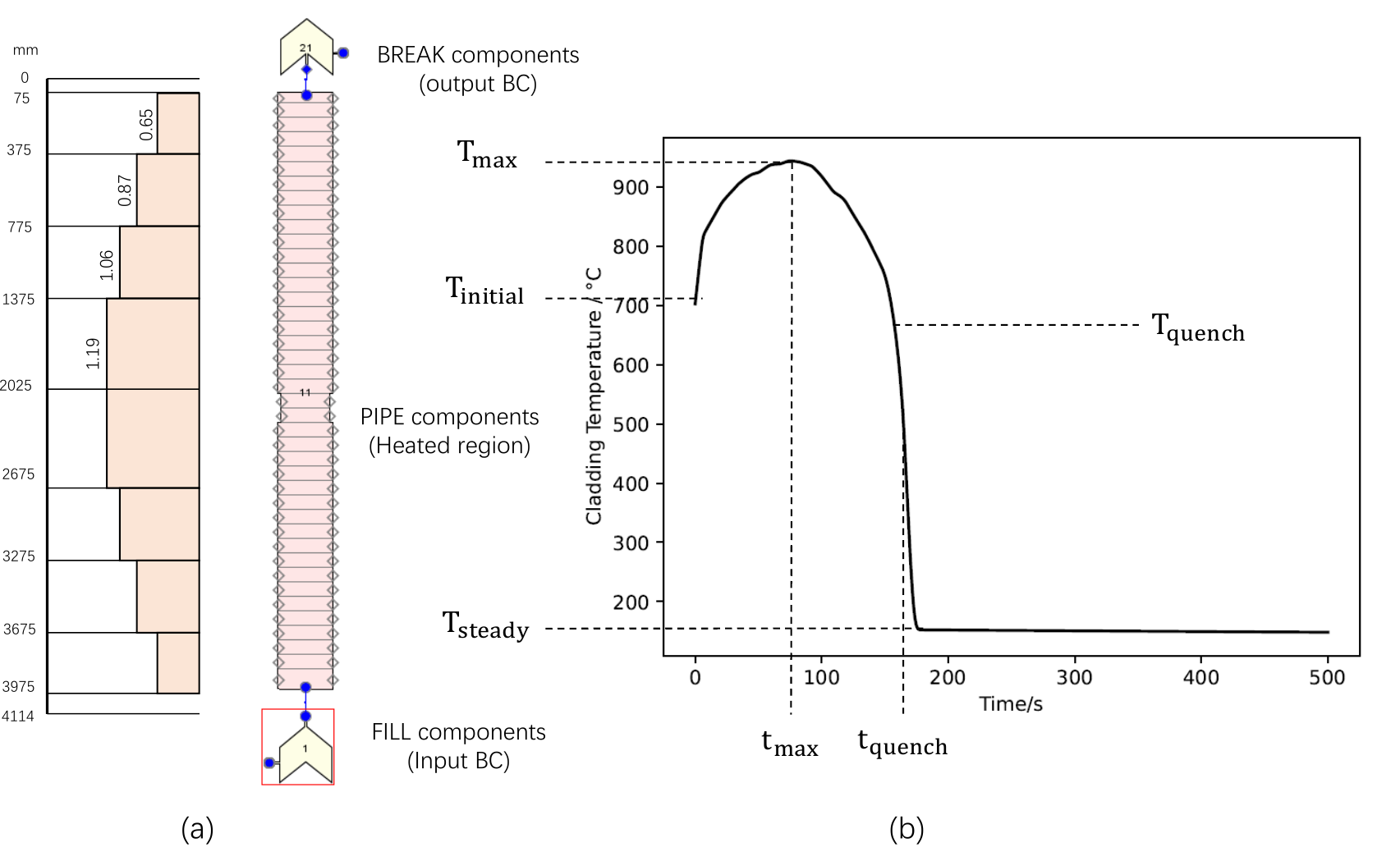}
	\caption[]{(a) TRACE simulation model and its related axial power profile for FEBA test 216; (b) A typical PCT profile from TRACE simulation.}
	\label{figure:fig_FEBA_temperature_pattern}
\end{figure}

In the 1980s, the Karlsruhe Institute of Technology carried out a series of experiments known as FEBA \cite{ihle1984feba1} \cite{ihle1984feba2} to improve the understanding of heat transfer during reflooding. The experiment facility consisted of a full height $5 \times 5$ bundle of pressurized water reactor rod simulators, with a heater that provided a cosine power profile over the height of the rod, which is shown in Figure \ref{figure:fig_FEBA_temperature_pattern} (a). The length of the rod was 4 m, with a heated length of 3.9 m, and the cladding temperature was measured at eight different elevations. This paper focuses on FEBA test series 1 test number 216, which is the baseline test with no flow  blockage and undisturbed bundle geometry containing all grid spacers. The experimental conditions are: water flooding velocity of 3.8 cm/s, system pressure of 4.1 bars, feedwater temperatures of 48\textdegree C in the first 30s and 37\textdegree C at the end respectively, and power starting at 200 kW and decay heat transient corresponding 120\% of ANS Standard about 40 seconds after reactor shutdown \cite{ihle1984feba2}. The reason for selecting this test is that it was well studied in the PREMIUM benchmark \cite{reventos2016premium}. The TRACE (v5.0p5) \cite{USNRC2014TRACE} system TH code is used to simulate the experiments based on the given initial and boundary conditions in FEBA test 216. Figure \ref{figure:fig_FEBA_temperature_pattern} shows the model built for TRACE simulation in this work and a typical peak cladding temperature (PCT) time series profile from TRACE simulation.

For FEBA test 216, the time-dependent PCTs were measured at 8 different axial positions over the bundle. In this specific problem, we choose the experimental data at axial position $z = 2225$ mm for IUQ and data at other 2 axial positions for validation ($z = 1135$ mm and $z = 3315$ mm). The QoI is the whole transient PCT profile, which contains major phase and magnitude information including the maximum PCT ($T_{\text{max}}$), time to reach the maximum PCT ($t_{\text{max}}$), and the time of quenching ($t_{\text{quench}}$).

In the TRACE model of the FEBA experiment, 36 uncertain physical model parameters in UQ section of TRACE system code \cite{nrc2014trac} were initially considered. However, not all of these parameters are significant to the QoIs. To reduce the input dimension by identifying the significant physical model parameters, our previous research \cite{xie2022bayesian-physor} performed a global sensitivity analysis study, resulting in the selection of four physical model parameters that are significant to the QoIs. These parameters are multiplicative factors that can be perturbed in the TRACE input deck, and their nominal values are 1.0, as shown in Table \ref{table:list_of_parameters_and_priors}. These four physical model parameters will be considered as calibration parameters in the IUQ study. Uninformative uniform distributions were chosen for the priors in the range of $[0,5]$. The \textit{objective of IUQ} is to determine the posterior distributions of these calibration parameters based on the chosen experimental data, such that the agreement between TRACE simulation and the FEBA experimental data can be improved. Furthermore, the quantified posterior uncertainties in these physical model parameters are expected to result in better TRACE prediction of experimental tests whose data is not used in the IUQ process.

\begin{table}[!htb]
	\footnotesize
	\caption{Selected TRACE physical model parameters for IUQ.}
	\label{table:list_of_parameters_and_priors}
	\centering
	\begin{tabular}{l c c c}
		\toprule
		Calibration parameters $\bm{\theta}$ (multiplication factors) & Representations & Uniform ranges  & Nominal \\ 
		\midrule
		Single phase vapor to wall heat transfer coefficient 			& \texttt{P1009} & (0.0, 5.0) & 1.0 \\
		Film to transition boiling Tmin criterion temperature 			& \texttt{P1010} & (0.0, 5.0) & 1.0 \\
		Dispersed flow film boiling heat transfer coefficient 			& \texttt{P1011} & (0.0, 5.0) & 1.0 \\
		Interfacial drag (dispersed flow film boiling) coefficient      & \texttt{P1031} & (0.0, 5.0) & 1.0 \\
		\bottomrule
	\end{tabular}
\end{table}

\section{Methodologies}
\label{section:Methods}

\subsection{Principal Component Analysis}
\label{section:Methods-conventional-PCA}

Since the QoIs in this problem are time-dependent, which corresponds to an infinite-dimension response, one may pick the PCT values at many time points to adequately represent the time series evolution. This, however, will result in high-dimensional outputs that are also highly correlated. Because a surrogate model for TRACE has to be used in order to reduce the computational cost in MCMC sampling, it is impractical and computationally expensive to create separate surrogate models for all the outputs. To address this challenge, PCA, an unsupervised ML method, is usually employed to reduce the dimensionality of high-dimensional correlated data. PCA is a statistical procedure that uses an orthogonal transformation to convert possibly correlated data into a set of linearly uncorrelated variables. The resulting PCs are orthogonal to each other and the corresponding PC scores are treated as values of new variables, whose number is much smaller than the number of original variables. Furthermore, the limited number of selected PCs can still preserve the majority of the original data variance after dimensionality reduction.

Once TRACE is used to simulate the time-dependent PCT profile, $p = 1000$ points are chosen evenly from the PCT profile. Note that a series of numerical tests has shown that such a number of points is sufficient, as it gives the same PCA results with cases when much larger values for $p$ are used. Next, $N = 500$ samples are generated from prior distribution of input parameters by Latin-hypercube sampling (LHS), which is listed in Table \ref{table:list_of_parameters_and_priors}. This results in a $p\times N$ data matrix $\mathbf{A}$, in which the rows represent the high-dimensional correlated outputs and the columns represent different samples. To transform the original data matrix $\mathbf{A}$ into an uncorrelated set of variables, we seek to find a $p\times p$ linear transform matrix $\mathbf{P}$. The linear transformation $\mathbf{PA}=\mathbf{B}$ will result in a new $p\times N$ data matrix $\mathbf{B}$, which contains the samples of the transformed uncorrelated variables. A typical 2-dimension PCA process is shown in Figure \ref{figure:PCA_example}. To find the matrix $\mathbf{P}$, we use the following steps:

\begin{figure}[htp!]
\centering
\includegraphics[width=0.65\textwidth]{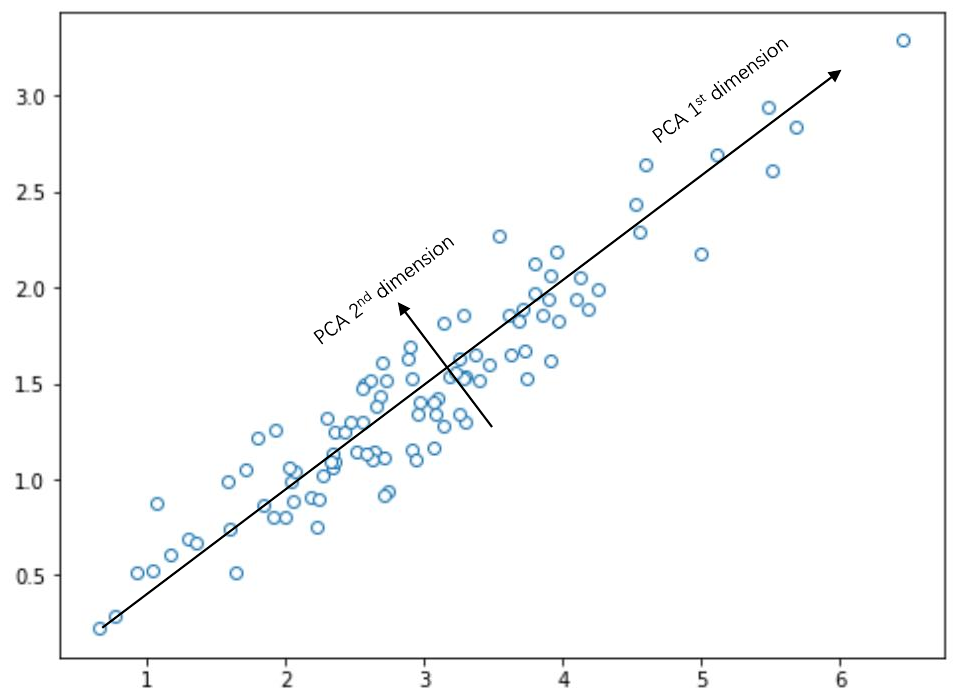}
\caption{PCA of a multivariate Gaussian distribution centered at (3,1.5). The vector shows the new base based on the PCA result.}
\label{figure:PCA_example}
\end{figure}

\begin{enumerate} \setlength{\itemsep}{0pt}
	\item Center the original data matrix $\mathbf{A}$ by defining the row means as a column vector $\mathbf{u}$. The centered data matrix $\mathbf{A}_{\text{centered}}$ is obtained by subtracting $\mathbf{u}$ form each column of $\mathbf{A}$.
	
	\item Find the singular value decomposition (SVD) of $\mathbf{A}_{\text{centered}}$:
	\begin{equation}
		\mathbf{A}_{\text{centered}} = \mathbf{U} \mathbf{\Lambda} \mathbf{V}^{\top}
	\end{equation}
	Where $\mathbf{U}$ is a $p\times p$ orthogonal matrix , $\mathbf{V}$ is a $N\times N$ orthogonal matrix, and $\mathbf{\Lambda}$ is a $p\times N$ diagonal matrix with non-negative real numbers on the diagonal. The diagonal entries of $\mathbf{\Lambda}$ are called the singular values of $\mathbf{A}_{\text{centered}}$ and are arranged in descending order.
	
	\item Choose $\mathbf{P} = \mathbf{U}^{\top}$, then we have:
	\begin{equation}
		\mathbf{PA}_{\text{centered}} = \mathbf{U}^{\top}\mathbf{A}_{\text{centered}} = \mathbf{\Lambda} \mathbf{V}^{\top} = \mathbf{B}
	\end{equation}
	In this case, it can be proven that the new variables in the new data matrix $\mathbf{B}$ are uncorrelated because its covariance matrix is diagonal. The matrix $\mathbf{P}$ provides a linear transformation from the original data basis to the PC basis. The rows of $\mathbf{P}$ are the PCs. The columns of $\mathbf{B}$ contain the samples of the transformed variables, also called PC scores.
	
	\item Determine the reduced dimension of the PC subspace $p^{*}$ which could be much smaller than $p$ based on the total variance explained by the PC subspace, using the diagonal entries in $\mathbf{\Lambda}$. Usually, the variances explained by the PCs decrease rapidly, only a few PCs can explain 95\% to 99\% of the total variance. Using a small value of $p^{*}$, define a $p^{*} \times p$ transformation matrix $\mathbf{P^{*}}$
	\begin{equation}
		\mathbf{P^{*}A}_{\text{centered}} = \mathbf{B^{*}}
	\end{equation}
	where $\mathbf{B^{*}}$ is a new data matrix with low-dimensional uncorrelated variables.
	
	\item To reconstruct the original PCT time series profile based on a sample $\mathbf{b}^{*}$ in the PC subspace, we use the following relation:
	\begin{equation}
		\mathbf{a}_\text{centered} = (\mathbf{P}^{*})^{\top}\mathbf{b}^{*}
	\end{equation}
	Then the mean vector $\mathbf{u}$ computed in step 1 will be added to $\mathbf{a}_\text{centered}$, to obtain the original data series profile.
\end{enumerate}

Through this PCA process, we can reduce the dimension of the QoIs from $p = 1000$ (or even more, depending on how many points are picked from the transient curve) to less than 10. If the selected PCs are used as QoIs in IUQ process, the experimental data also need to be transferred into the PC subspace. The uncertainty of experimental data also needs to be transformed in a similar way using the following relation:
\begin{equation}    \label{equation:PCA-Error}
	\mathbf{\Sigma}^{*}_{\text{data}} = \mathbf{P^{*}}\mathbf{\Sigma}_{\text{data}}(\mathbf{P^{*}})^\top
\end{equation}
where $\mathbf{\Sigma}_{\text{data}}$ is a $p\times p$ matrix that includes the uncertainty of experimental data. It can be a full matrix if the correlations between the high-dimensional correlated responses are known. However, such information is usually not available so one may assume $\mathbf{\Sigma}_{\text{data}}$ is a diagonal matrix. From equation (\ref{equation:PCA-Error}) we can find that the new variance $\mathbf{\Sigma}^{*}_{\text{data}}$ in the PC subspace is usually a $p^{*}\times p^{*}$ full matrix with non-zero off-diagonal entries. This new data uncertainty matrix needs to be considered in the Bayesian IUQ process.

\subsection{Functional PCA}
\label{section:Methods-fPCA}

\begin{figure}[htp!]
	\centering
	\includegraphics[width=0.95\textwidth]{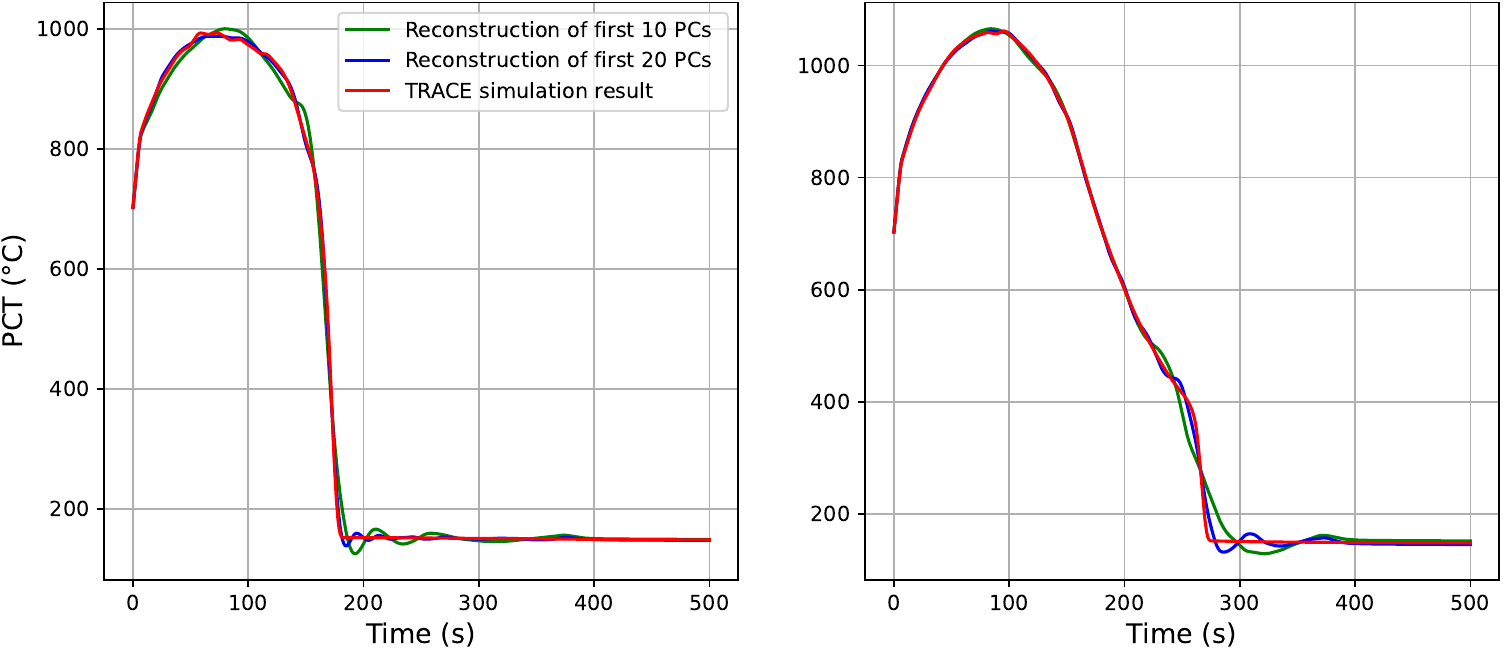}
	\caption{Illustration of the inaccuracies in the reconstructed PCT profiles using conventional PCA for 2 simulation tests with different physical model parameters.}
	\label{figure:PCA_reconstruction_inaccuracy}
\end{figure}

In the conventional PCA method described above, the original data matrix is centered in the first step, and the mean vector has to be added back in order to reconstruct the data. As shown in Figure \ref{figure:fig_FEBA_temperature_pattern}, each TRACE simulated PCT profile has its own phase and magnitude information, $t_{\text{max}}$, $t_{\text{quench}}$, and $T_{\text{max}}$. Using the mean vector will ``smooth out'' such important information. As a result, the conventional PCA method may not be able to recover such phase and magnitude information accurately using only first few PCs, even though they explain more than 99\% of the total variance. There may be non-negligible fluctuations in the reconstructed PCT profiles near the quenching time, which is shown in Figure \ref{figure:PCA_reconstruction_inaccuracy}. To solve this problem, functional alignment \cite{wang1997alignment} \cite{ramsay1998curve} is used to separate the phase and magnitude information of the original data matrix before dimensionality reduction. The combination of functional alignment and conventional PCA will be referred to as \textit{functional PCA (fPCA)} in the following.

Functional alignment aims in aligning the ``landmark'' points, which are $t_{\text{max}}$ and $t_{\text{quench}}$ in this problem, of the whole dataset to the same points. A composite function $\Tilde{f}(t) = f(\gamma(t))$ is used to adjust the original function, where $\gamma(t)$ is called the \textit{warping function}, and $\Tilde{f}(t)$ is the \textit{warped data}. The set of all warping functions $\gamma(t)$ ($\Gamma$) will have the following property:
\begin{equation}
	\Gamma = \{\gamma : [0,t]\rightarrow [0,t] \mid \gamma(0) = 0, \gamma(t) = t, \gamma\text{ is a monotonically increasing function} \}
\end{equation}

The main problem is to find the warping functions $\gamma(t)$ that can align all the functions at the landmark points. Many methods have been developed for determining $\gamma(t)$ through minimizing the cost function $\inf _{\gamma \in \Gamma} ||f_1(t) - f_2(\gamma(t))||$ \cite{wang1997alignment} \cite{ramsay1998curve}. Here, we will introduce the square root slope function (SRSF) method \cite{tucker2013generative} \cite{tucker2014functional}, which uses the square root slope to represent the original function. The SRSF of the original function $f(t)$ is defined in the following form: 
\begin{equation}
	q(t) = \mathrm{sign}(\Dot{f}(t))\sqrt{\lvert\Dot{f}(t)\rvert}
\end{equation}

If $f(t)$ is a continuous function, then the SRSF $q(t)$ is square-integrable. The function $f(t)$ can be calculated using the integral $f(t) = f(0)+\int_0^t q(s)\lvert q(s)\rvert ds$, since $q(s)\lvert q(s)\rvert = \dot{f}(s)$. If we warp a function $f$ by $\gamma$, the SRSF of $f(\gamma(t))$ is given by:
\begin{equation}
	\Tilde{q}(t) = q(\gamma(t))\sqrt{\Dot{\gamma(t)}}
\end{equation}

A new cost function is defined based on the norm of two SRSFs. The warping function $\gamma(t)$ is determined by minimizing this cost function.
\begin{equation}
	D_{y}\left(f_{1}, f_{2}\right)=\inf _{\gamma \in \Gamma}\left\|q_{1}-\left(q_{2}(\gamma)\right) \sqrt{\dot{\gamma}}\right\|
\end{equation}

\begin{figure}[!htb]
	\centering
	\includegraphics[width=0.99\textwidth]{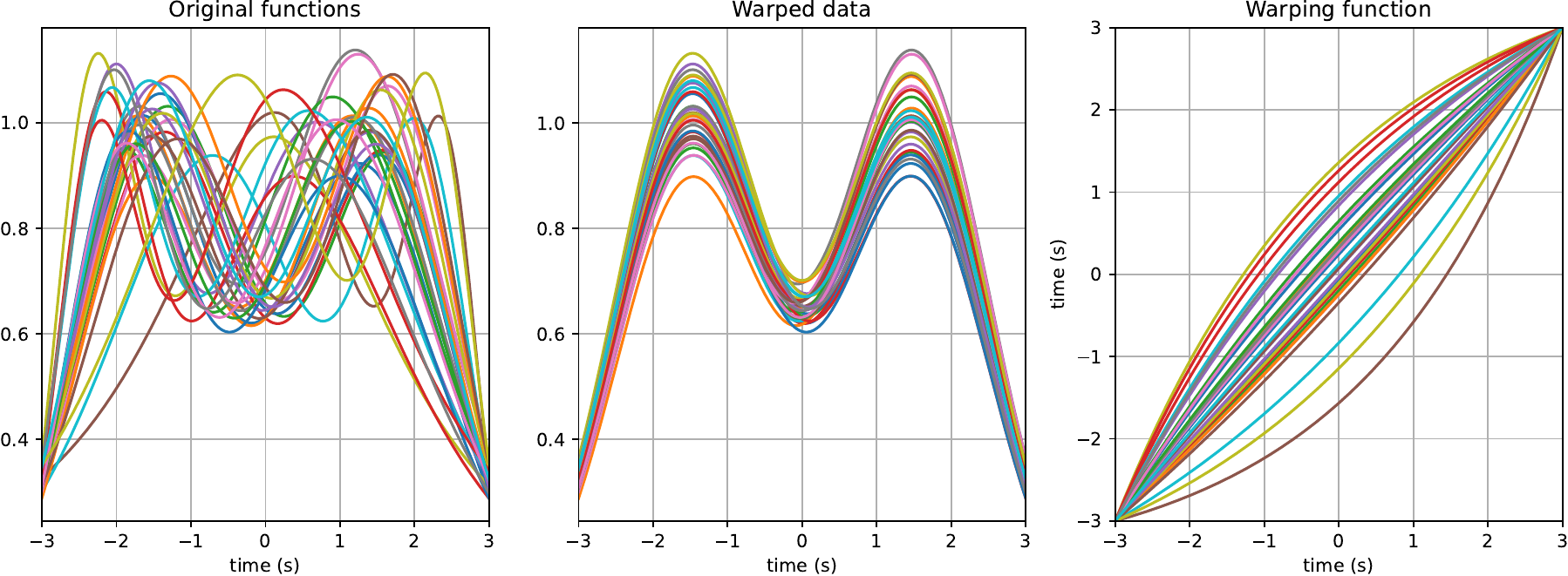}
	\caption[]{Illustration of functional alignment with some simple test functions: left: curves with different phases and amplitudes; middle: warped data that have curves aligned at the crest and trough points; right: warping functions.}
	\label{figure:fig2_fPCA_warping_example}
\end{figure}

After the separation of amplitude and phase information for all of the samples, the original samples are transferred into a series of warped data with aligned landmarks points and warping function includes phase information. Figure \ref{figure:fig2_fPCA_warping_example} shows an example of functional alignment of a series of functions. Afterwards, conventional PCA is applied to all warped data $\Tilde{f}(t)$ and warping functions $\gamma(t)$ for dimensionality reduction. The data of $\Tilde{f}(t)$ and $\gamma(t)$ could be represented by the first few PCs. To reconstruct the original function $f(t)$ using the limited number of PCs, warped data $\Tilde{f}(t)$ and warping function $\gamma(t)$ are first reconstructed based on the related PCs based on inverse PCA.

Finally, phase and amplitude reconstructed functions are combined through $f(t) = \Tilde{f}(\gamma^{-1}(t))$ to generate the original function before functional alignment. Since the warping function should be monotonically increasing, a smoothing function is applied to the PCA reconstructed $\gamma(t)$ functions to avoid non-monotonic issues calculating the inverse function $\gamma^{-1}(t)$. Note that previously curve registration and alignment for FEBA benchmark has been applied in an earlier work \cite{perret2019global}. The major focus of this work is building DNN-based surrogate models for the PC scores after functional alignment and its application in IUQ and FUQ.

\begin{figure}[!htb]
	\centering
	\includegraphics[width=0.99\textwidth]{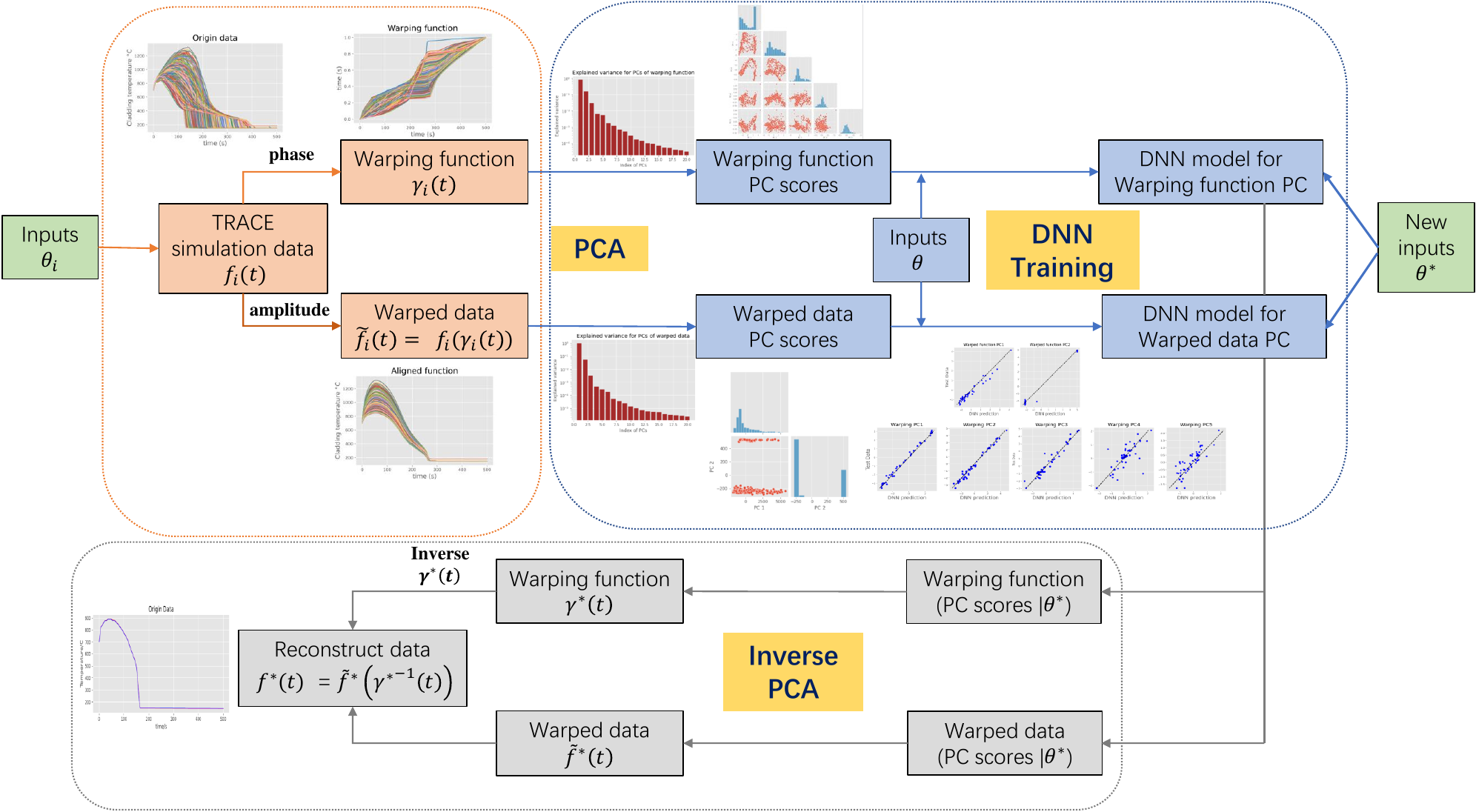}
	\caption[]{Procedure for fPCA and its inverse process. It includes the following major steps: (1) run TRACE at different samples of $\bm{\theta}$ to generate the PCT profiles, (2) apply functional alignment of the PCT profiles to obtain the warped data (aligned at $t_{\text{max}}$ and $t_{\text{quench}}$) and warping functions, (3) apply PCA for dimensionality reduction of the warped data and warping functions, (4) build and validate the surrogate models based on the resulting PC scores (samples of the transformed new variables), (5) at a new sample $\bm{\theta}^{*}$, run the surrogate models to get the estimated PC scores, (6) use reverse PCA to obtain the corresponding warped data and warping function, (7) use reverse functional alignment to obtain the original PCT profile.}
	\label{figure:fig_fpca_flowchart}
\end{figure}

Figure \ref{figure:fig_fpca_flowchart} shows the procedure of fPCA application. In this framework, surrogate models like DNN are used to represent the PC scores from phase and amplitude information, respectively. When new samples are given, the predictions of DNN surrogate models go through an ``inverse fPCA'' process, as discussed above, to reconstruct the original time series data.

\subsection{Bayesian Neural Networks} 
\label{section:Methods-BNN}

\begin{figure}[!htb]
	\centering
	\includegraphics[width=0.8\textwidth]{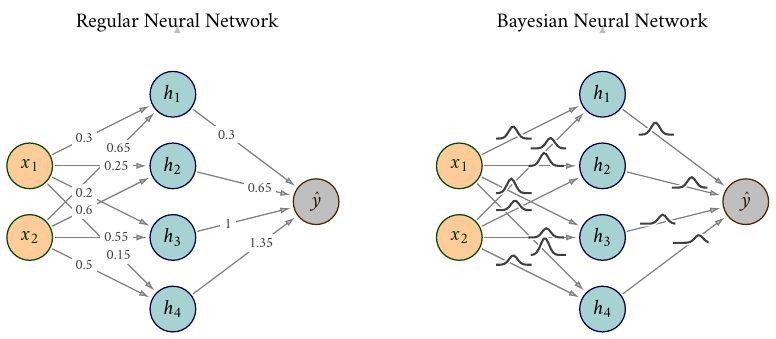}
	\caption[]{An illustration of the uncertain parameters (only weights are shown) in a BNN compared to a standard neural network that uses scalar weights.}
	\label{figure:UQ-DNN-BNN}
\end{figure}

In a standard neural network structure, the learnable parameters (weights and biases) are initialized randomly with deterministic values. During the prediction stage, one would anticipate a deterministic output for a given input since the weights and biases have fixed values after training. In contrast, a BNN \cite{ghahramani2016history} \cite{goan2020bayesian} is a neural network in which the learnable parameters follow random distributions. To train a BNN, prior distributions are assigned to the neural network parameters, and then, based on the training data, posterior distributions of the parameters are computed by updating the prior distributions during training process. Figure \ref{figure:UQ-DNN-BNN} compares a standard neural network with a BNN. Following training, the BNN is evaluated at the same input several times, each time with its parameters sampled from the posterior distributions, resulting in different values for the prediction that can be used to obtain the predictive uncertainties.

The inference of the posterior distributions is challenging since most DNNs nowadays have large number of parameters. To address this issue, various methods have been developed for Bayesian inference of neural networks, including sampling-based methods such as MCMC \cite{neal2012bayesian} and optimization-based methods like variational inference \cite{blei2017variational} \cite{tzikas2008variational}. Variational methods are advantageous because they converge faster, making them more suitable for large neural networks. In this study, we used variational inference to train the BNN. Note that we have treated the bias parameters as deterministic, as neural network predictions are less sensitive to these parameters than to weights. This is because the bias term is added to the product of weights and the activation from the previous hidden layer, and thus the impact of weights on DNN predictions is more significant than that of bias.

A probabilistic model is assumed for the BNN, in which the weights are learned by using Maximum Likelihood Estimation (MLE). The posterior weights $(\mathbf{w})$ are computed during training based on Bayes' rule for a given training dataset $(\mathcal{D})$:
\begin{equation}
	P (\mathbf{w} | \mathcal{D})  =  \frac{P (\mathcal{D} | \mathbf{w}) \cdot P (\mathbf{w})}{P (\mathcal{D})}
\end{equation}
where $P (\mathbf{w})$ is the prior distribution for $\mathbf{w}$ and it is assumed to be certain non-informative distribution, $P (\mathcal{D} | \mathbf{w})$ is the likelihood function, and $P (\mathbf{w} | \mathcal{D})$ is the posterior distribution for $\mathbf{w}$. Prior and posterior represent our knowledge of $\mathbf{w}$ before and after observing $\mathcal{D}$, respectively. $P (\mathcal{D})$ does not contain $\mathbf{w}$ so it is usually treated as a normalizing constant. It is sometimes referred to as the evidence term. When making predictions at a test data $\mathbf{x}^{*}$, the predictive distribution of the output $\mathbf{y}^{*}$ is given by:
\begin{equation}
	P (\mathbf{y}^{*} | \mathbf{x}^{*})  =  \mathbb{E}_{P (\mathbf{w} | \mathcal{D})} \left[  P (\mathbf{y}^{*} | \mathbf{x}^{*}, \mathbf{w})  \right]
\end{equation}
where the expectation operator $\mathbb{E}_{P (\mathbf{w} | \mathcal{D})}$ means we need to integrate over $P (\mathbf{w} | \mathcal{D})$. The term $ P (\mathbf{y}^{*} | \mathbf{x}^{*}, \mathbf{w})$ represents the probability of the prediction at a test point $\mathbf{x}^{*}$ and the posteriors of the weights. Each possible configuration of the weights, weighted according to the posterior distribution $P (\mathbf{w} | \mathcal{D})$, makes a prediction about $\mathbf{y}^{*}$ given $\mathbf{x}^{*}$. This is why taking an expectation under the posterior distribution on weights is equivalent to using an ensemble of an infinite number of neural networks. Unfortunately, such expectation operation is intractable for neural networks of any practical size, due to a large number of parameters as well as the difficulty to perform exact integration. This is the main motivation to use a variational approximation for $P (\mathbf{w} | \mathcal{D})$. Variational inference methods are a family of techniques for approximating intractable integrals arising in Bayesian inference and ML. It is used to approximate complex posterior probabilities that are difficult to evaluate directly as an alternative strategy to MCMC sampling. An alternative variational distribution is proposed to approximate $P (\mathbf{w} | \mathcal{D})$, it consists of a distribution set whose parameters are optimized using the Kullback-Leibler divergence. For more mathematical and implementation details, interested readers are recommended to look at \cite{yaseen2023quantification} \cite{blei2017variational} \cite{tzikas2008variational}.

\subsection{Bayesian Framework for IUQ}
\label{section:Methods-IUQ}

FUQ requires knowledge of computer model input uncertainties to generate the uncertainty of simulation outputs. These inputs uncertainties are often determined by ``expert opinion'' or ``user self-evaluation''. However, such determination lacks mathematical rigor and may be subjective, which may lead to misleading FUQ results. IUQ is a method to inversely quantify the input uncertainties based on the given experiment data while keeping the simulation results consistent with the experiment data. A modular Bayesian framework for IUQ has been developed \cite{wu2018inversePart1} previously and it will be used in this work. In the following we will provide a brief introduction to Bayesian IUQ.

Consider a computer model $\mathbf{y}^{\text{M}} \left( \mathbf{x}, \bm{\theta} \right)$, where $\mathbf{y}^{\text{M}}$ is the model response, $\mathbf{x}$ is the vector of design variables, and $\bm{\theta}$ is the vector of calibration parameters. The differences of design and calibration variables have been discussed in our previous work \cite{wu2021comprehensive}. Given the design variable $\mathbf{x}$, the reality $\mathbf{y}^{\text{R}}\left( \mathbf{x} \right)$ can be learned by (1) running model simulation, which involves model uncertainty $\delta(\mathbf{x})$, and (2) performing experiments, which involves measurement uncertainty $\bm{\epsilon}$. These terms can be combined in the so-called ``model updating equation'' \cite{kennedy2001bayesian}:
\begin{equation}	 \label{equation:IUQ2-Model-Update-Eqn}
	\mathbf{y}^{\text{E}} (\mathbf{x}) = \mathbf{y}^{\text{M}} \left( \mathbf{x}, \bm{\theta}^{*} \right) + \delta(\mathbf{x}) + \bm{\epsilon}
\end{equation}
where $\delta(\mathbf{x})$ is the model uncertainty/discrepancy, due to missing/incomplete physics and numerical approximation errors during the modeling process. $\bm{\theta}^{*}$ is the ``true" but unknown values of $\bm{\theta}$. $\bm{\epsilon} \sim \mathcal{N} \left( 0, \Sigma_{\text{exp}} \right)$ represents the measurement/experimental uncertainty which is considered as a normal distribution. The model discrepancy term $\delta(\mathbf{x})$ is dependent on $\mathbf{x}$ which stands for design variables such as initial conditions or boundary conditions. Since only one experimental test is considered in this transient problem, we do not have enough cases of $\mathbf{x}$ to learn $\delta(\mathbf{x})$. Therefore, the model discrepancy is not considered in this study.

Based on the assumption that the experimental uncertainty is Gaussian, $\bm{\epsilon} = \mathbf{y}^{\text{E}} (\mathbf{x}) - \mathbf{y}^{\text{M}} \left( \mathbf{x}, \bm{\theta}^{*} \right)$ follows a multi-dimensional normal distribution. As a result, the posterior distribution $p \left( \bm{\theta}^{*} | \mathbf{y}^{\text{E}}, \mathbf{y}^{\text{M}}\right)$ can be written as:
\begin{equation} \label{equation:IUQ3-Posterior}
	\begin{split}
		p \left( \bm{\theta}^{*} | \mathbf{y}^{\text{E}}, \mathbf{y}^{\text{M}}\right) & \propto  p \left( \bm{\theta}^{*} \right) \cdot p \left( \mathbf{y}^{\text{E}}, \mathbf{y}^{\text{M}} | \bm{\theta}^{*} \right) 	\\
		& \propto  p \left( \bm{\theta}^{*} \right) \cdot \frac{1}{\sqrt{|\bm{\Sigma}|}}  \cdot \text{exp} \left[  - \frac{1}{2} \left[ \mathbf{y}^{\text{E}} - \mathbf{y}^{\text{M}} \right]^\top \bm{\Sigma}^{-1} \left[ \mathbf{y}^{\text{E}} - \mathbf{y}^{\text{M}}\right] \right]
	\end{split}
\end{equation}
where $p \left( \bm{\theta}^{*} \right)$ is the prior distribution that can be provided by user evaluation or expert opinion. $p \left( \mathbf{y}^{\text{E}}, \mathbf{y}^{\text{M}} | \bm{\theta}^{*} \right)$ is the likelihood function. Prior and posterior distributions represent our knowledge of $\bm{\theta}$ before and after observation of measurement data, respectively. $\bm{\Sigma}$ is the covariance of the likelihood which consists of two parts:
\begin{equation}
	\bm{\Sigma} = \bm{\Sigma}_{\text{exp}} + \bm{\Sigma}_{\text{code}}
\end{equation}
where $\bm{\Sigma}_{\text{exp}}$ is the experimental uncertainty due to measurement error, and $\bm{\Sigma}_{\text{code}}$ is the code/interpolation uncertainty due to the use of surrogate models to reduce the computational cost. The term $\bm{\Sigma}_{\text{code}} = 0$ if the computer model is used directly in the IUQ process, instead of the surrogate models. Note that a component for model uncertainty/discrepancy should be included in $\bm{\Sigma}_{\text{exp}}$ if possible. As discussed above, due to the very limited amount of data,  it is not considered in this work. To calculate the posterior distribution, an adaptive MCMC algorithm \cite{andrieu2008tutorial} is used to generate samples following the probability densities of posterior distributions. To reduce the computational cost of MCMC sampling, surrogate models built with DNN are used to represent the simulation of computer models. In this project, we will compare GP and DNN as surrogate models to determine which approach lead to better IUQ results. Note that when we say ``DNN-based surrogate models'' with consideration of the code uncertainty $\bm{\Sigma}_{\text{code}}$, we essentially mean BNN. BNN is a special implementation of DNN that accounts for the prediction uncertainty.

\section{Results and Discussions}
\label{section:Results}

In this paper, we compare four different methods for IUQ with different dimensionality reduction and surrogate modeling approaches. Table \ref{table:list_of_IUQ_methods} lists the details of these methods. The conventional PCA process follows Section \ref{section:Methods-conventional-PCA}, while the fPCA process follows Section \ref{section:Methods-fPCA}. In Method 1, the combination of conventional PCA and GP serves as a reference solution. GP has a unique feature compared to other ML methods, which is that the mean square error (MSE), also called the variance of the prediction is directly available. Therefore, $\bm{\Sigma}_{\text{code}}$ can be easily included in the IUQ process. Compared to DNN, GP is mainly used for problems with low-dimensional features and smooth responses. The combination of fPCA and GP has been explored in \cite{perret2022global}. Since the main focus of this work is to demonstrate the applicability and benefits of fPCA with DNN while accounting for $\bm{\Sigma}_{\text{code}}$, we will not repeat this approach. Methods 2 and 3 use conventional DNNs as surrogate models, while Method 4 uses BNNs.

\begin{table}[!htb]
	\centering
	\footnotesize
	\caption{Details of the four IUQ methods.  }
	\label{table:list_of_IUQ_methods} 
	\begin{tabular}{c c c c}
		\toprule
		Methods  &  PCA  &  Surrogate models  &  $\bm{\Sigma}_{\text{code}}$ considered?  \\ 
		\midrule
		Method 1 & Conventional  &  GP   &  Yes  \\
		Method 2 & Conventional  &  DNN  &  No   \\
		Method 3 & fPCA          &  DNN  &  No   \\
		Method 4 & fPCA          &  BNN  &  Yes  \\
		\bottomrule
	\end{tabular}
\end{table}

This section is arranged as follows. Section \ref{section:Results1-fPCA} introduces the functional alignment results of TRACE simulation samples and gives a comparison of conventional PCA with fPCA. Section \ref{section:Results2-surrogate} presents the results for validation of surrogate modeling and the UQ process for BNN models. Section \ref{section:Results3-IUQ} presents the IUQ results, including the posterior distributions of the calibration parameters and their mean values and standard deviations. Section \ref{section:Results4-FUQ-validation} presents the validation results for the IUQ results, including FUQ for experimental tests at 3 different axial positions and comparison with model simulations based on the prior distributions and the experimental data.

\subsection{Results for fPCA}
\label{section:Results1-fPCA}

To train the fast-running and accurate surrogate models for TRACE, 500 random samples are generated based on the prior distributions of 4 calibration parameters using LHS. Next, the conventional and functional PCA methods are performed to obtain the PC scores for the samples, which will be used as the training data for the surrogate models. For conventional PCA, we apply PCA to the PCT profiles directly. For fPCA, functional alignment is applied first to the PCT profiles before the PCA process.

\begin{figure}[!htb]
	\centering
	\includegraphics[width=0.99\textwidth]{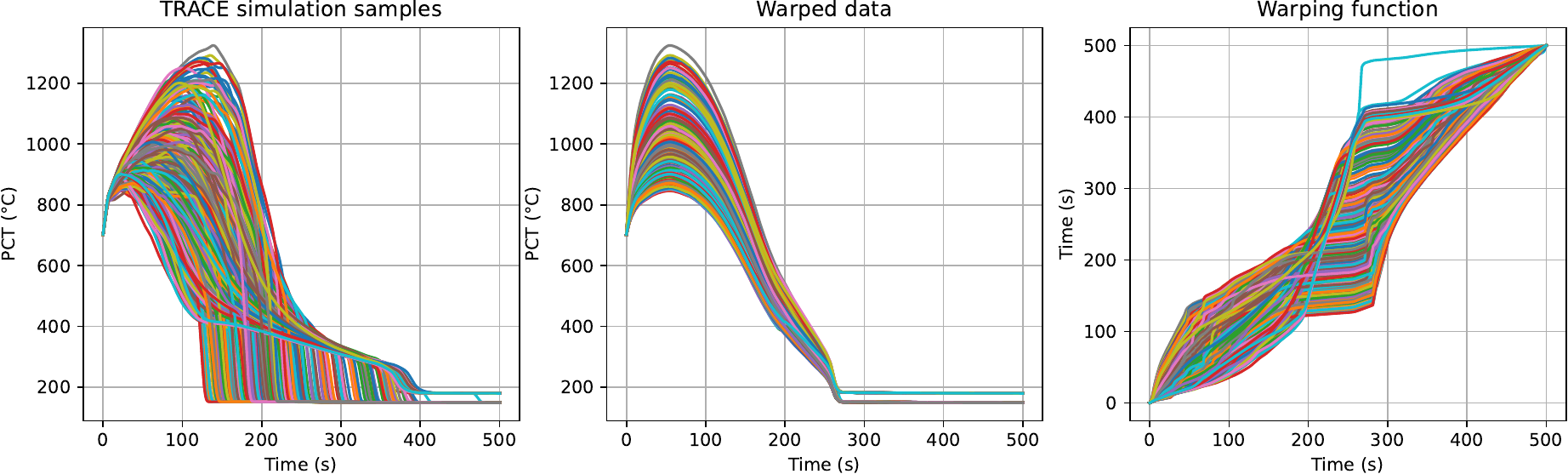}
	\caption[]{TRACE simulated PCT profiles before and after functional alignment. Left: original data; Middle: warped data (aligned at $t_{\text{max}}$ and $t_{\text{quench}}$); Right: warping functions.}
	\label{figure:fig_PCA_data}
\end{figure}

During fPCA, the original TRACE simulation data was separated into warped data which includes amplitude information, and warping function which contains phase information. The results are shown in Figure \ref{figure:fig_PCA_data}, the time to reach maximum PCT ($t_{\text{max}}$), and the time of quenching ($t_{\text{quench}}$) for all the 500 TRACE simulated PCT profiles are aligned to the same positions, respectively.

\begin{figure}[!htb]
	\centering
	\includegraphics[width=0.90\textwidth]{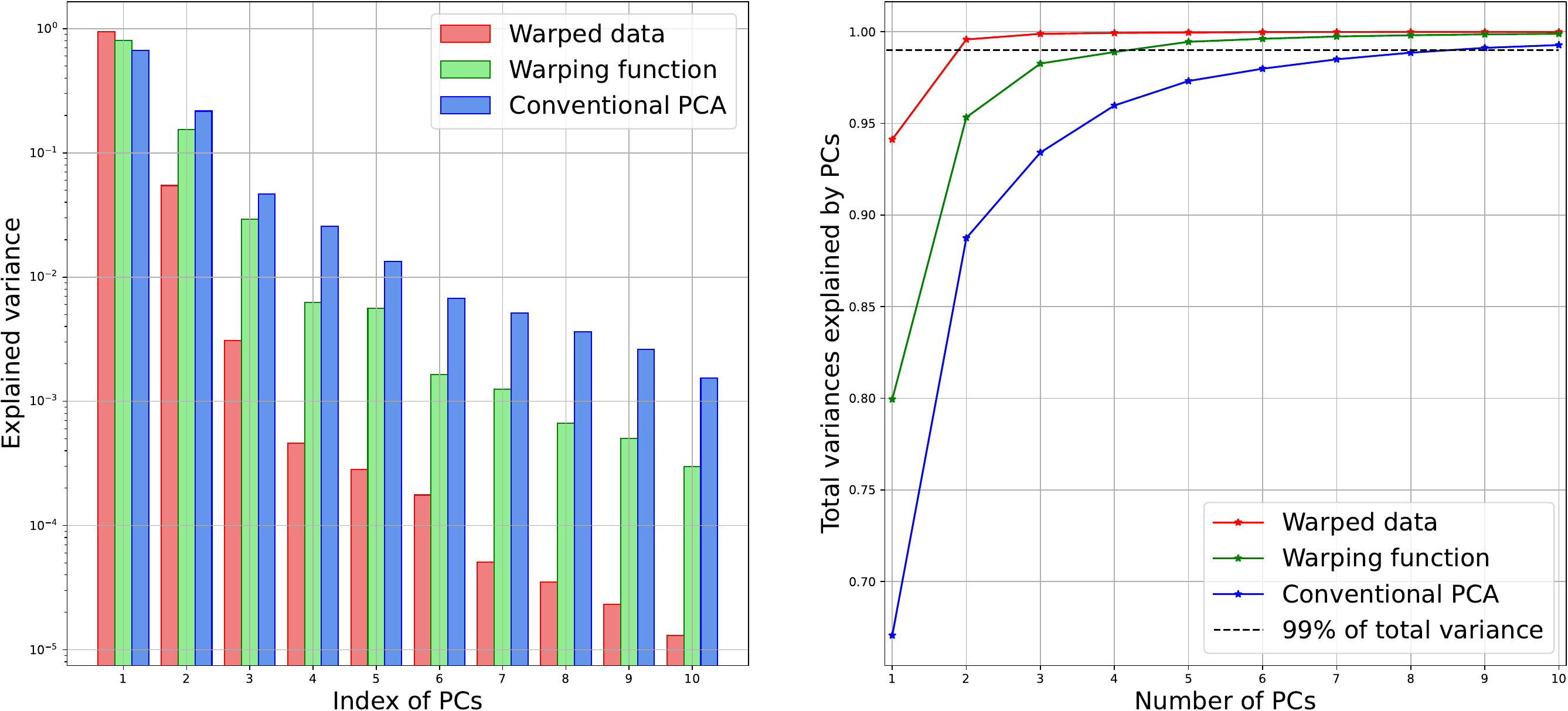}
	\caption[]{Left: Variance explained by each PC with and without functional alignment. Right: Total variance explained by first 10 PCs from PCA with and without functional alignment.}
	\label{figure:fig_PCA_variance}
\end{figure}

After functional alignment, we have two series of datasets, $\Tilde{f}(t)$ for warped data and $\gamma(t)$ for warping functions. Both datasets have 1000 time steps, which form two $1000 \times 500$ matrices, whereas the original dataset has only one $1000 \times 500$ data matrix. PCA is then applied to both datasets. Figure \ref{figure:fig_PCA_variance} shows the total variance explained by the first 10 PC scores for PCA of the warped data, the warping function, and the original TRACE simulation data without functional alignment. The fPCA process shows a significant improvement compared to the conventional PCA because we need fewer PCs to account for the 99\% of the total variance. The first 2 PCs for warped data and the first 4 PCs for warping functions are chosen as the new QoIs by fPCA, which can explain over 99\% of the total variance. For conventional PCA, we choose the first 4 PCs as QoIs, which explain 95\% of the total variance. The first 10 PCs would have to be chosen to account for 99\% of the total variance.

\begin{figure}[!htb]
	\centering
	\includegraphics[width=0.8\textwidth]{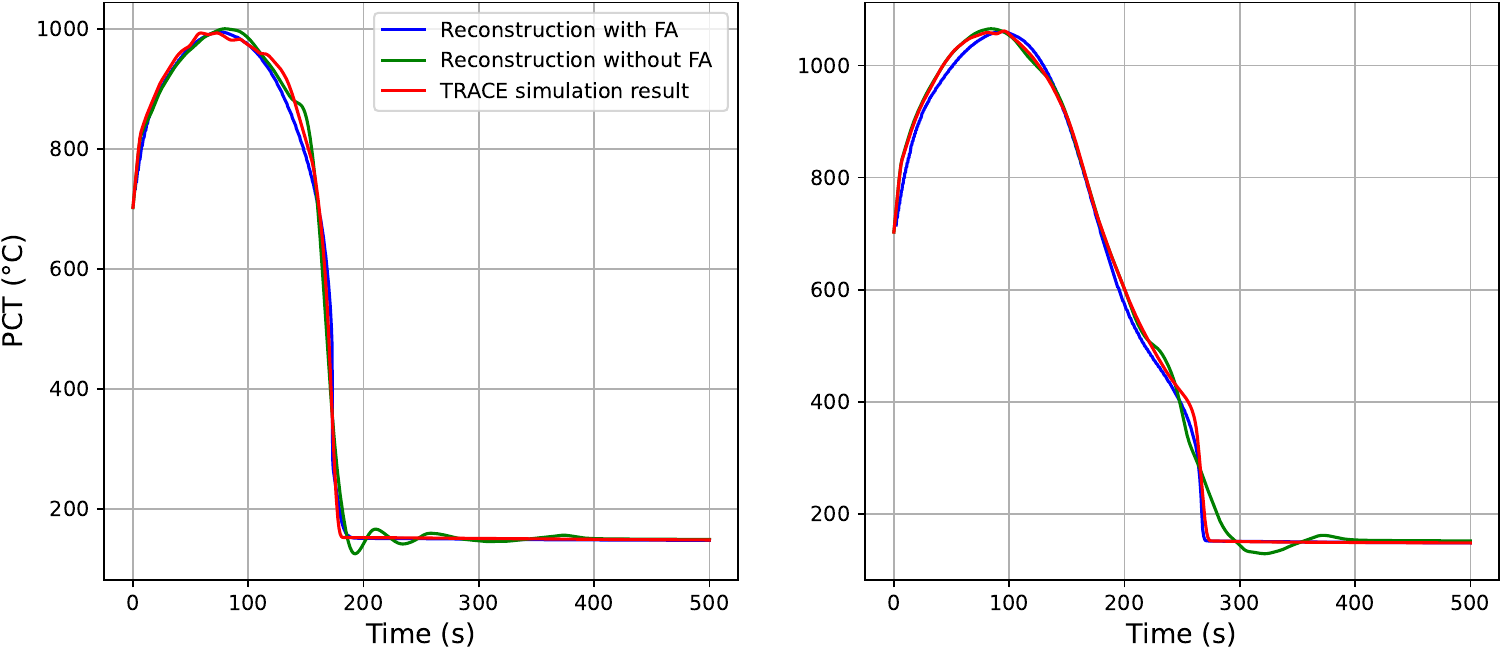}
	\caption[]{Comparisons of TRACE-simulated and reconstructed PCT profiles with and without functional alignment (FA in the figure), for 2 simulation tests with different physical model parameters.}
	\label{figure:fig_PCA_validation}
\end{figure}

Figure \ref{figure:fig_PCA_validation} shows the comparisons of TRACE simulation samples and reconstructed PCT profiles from PC scores with and without functional alignment. The reconstruction process means obtaining the original PCT profiles from the PC scores, as explained in Section \ref{section:Methods-conventional-PCA}. For the PCA method without functional alignment, we used 10 PCs, which explains the same variance as the PCs used in our fPCA study. In Figure \ref{figure:fig_PCA_validation}, the reconstructed PCT profiles based on 6 PCs from fPCA show a good agreement with the TRACE simulation results, while the reconstructed PCT profiles by conventional PCA show oscillations as expected, especially near the quenching time when the PCT has a sudden drop.

\subsection{Results for surrogate modeling}
\label{section:Results2-surrogate}

The QoIs, surrogate models and validation results of the four IUQ methods in Table \ref{table:list_of_IUQ_methods} are summarized in Table \ref{table:surrogate_model_details}. Before training the surrogate models, all the PC scores are standardized to zero mean and unit variance and separated into three groups, 70\% for training, 15\% for validation and 15\% for testing. All the surrogate models take the four calibration parameters in Table \ref{table:list_of_parameters_and_priors} as inputs. For Method 1, one multi-dimensional GP model was trained for 4 PCs as outputs. For methods 2-4, separate DNN/BNN models were used to represent each PC. Neural network models can certainly represent multi-dimensional responses. However, it was found that the accuracy was not as good as training separate DNNs/BNNs for each PC as response. One possible reason is that the PCs scores are essentially samples for transformed uncorrelated variables. When they are used to train a single DNN/BNN, all the layers/neurons before the output layer are shared, while only the weights from the last hidden layer to the output layer are different. This can cause the DNN/BNN to have a less satisfactory performance. We have also performed hyperparameter tuning with grid search to find optimized neural network architectures, learning rates, etc. Because the amount of training data is small for this simple problem, the DNN models only have 3 hidden layers with 10, 20, and 10 hidden neurons, and 1 output layer with one neuron to represent the PC.

\begin{table}[!htb]
	\centering
	\footnotesize
	\caption{QoIs, surrogate models and validation results for different IUQ methods.}
	\label{table:surrogate_model_details} 
	\begin{tabular}{l c c c c}
		\toprule
		Methods & Method 1 & Method 2 & Method 3 & Method 4  \\ 
		\midrule
		QoIs & \multicolumn{2}{c}{4 PCs (conventional PCA)} & \multicolumn{2}{c}{2 PCs + 4 PCs (fPCA)} \\
		Surrogate models & one GP & four DNNs & six DNNs & six BNNs\\
		Validation results & Figure \ref{figure:fig_surrogate_Validation_method1} & Figure \ref{figure:fig_surrogate_Validation_method2} & Figure \ref{figure:fig_surrogate_Validation_method3} & Figure \ref{figure:fig_surrogate_Validation_method4}\\
		\bottomrule
	\end{tabular}
\end{table}

\begin{figure}[!htb]
	\centering
	\includegraphics[width=0.99\textwidth]{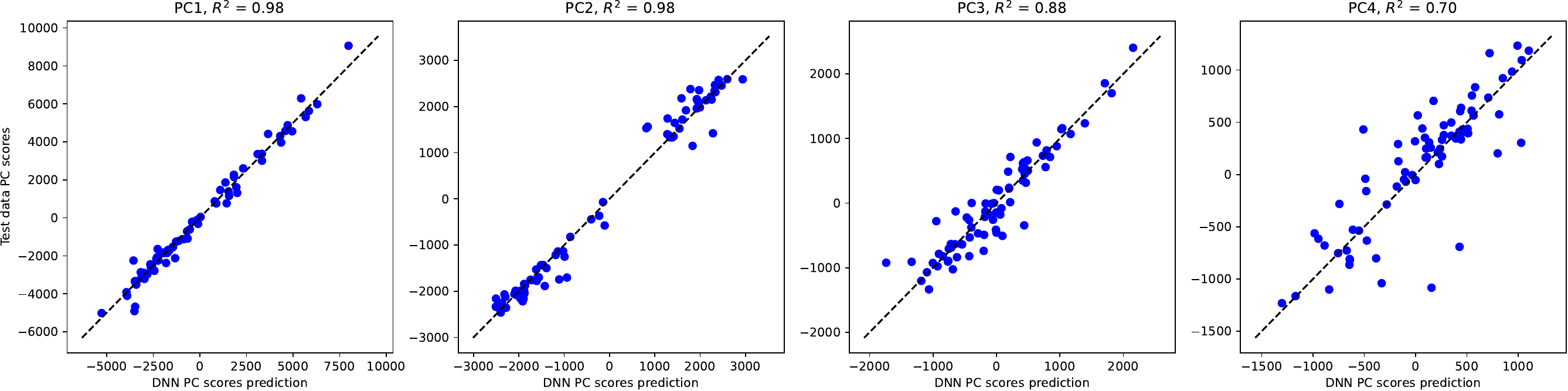}
	\caption[]{Validation of the GP surrogate model predictions for Method 1.}
	\label{figure:fig_surrogate_Validation_method1}
\end{figure}

\begin{figure}[!htb]
	\centering
	\includegraphics[width=0.99\textwidth]{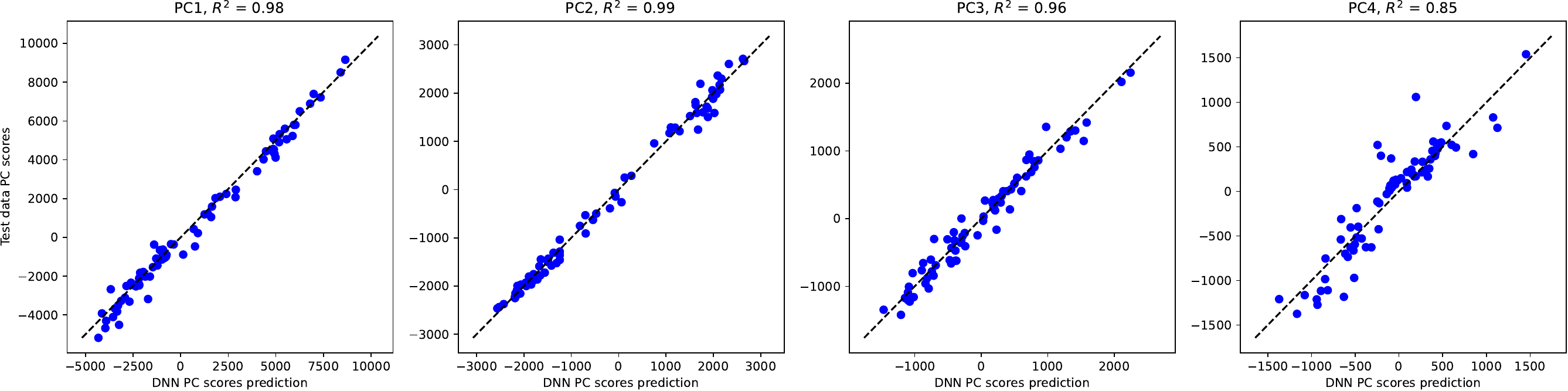}
	\caption[]{Validation of the DNN surrogate model predictions for Method 2.}
	\label{figure:fig_surrogate_Validation_method2}
\end{figure}

\begin{figure}[!htb]
	\centering
	\includegraphics[width=0.8\textwidth]{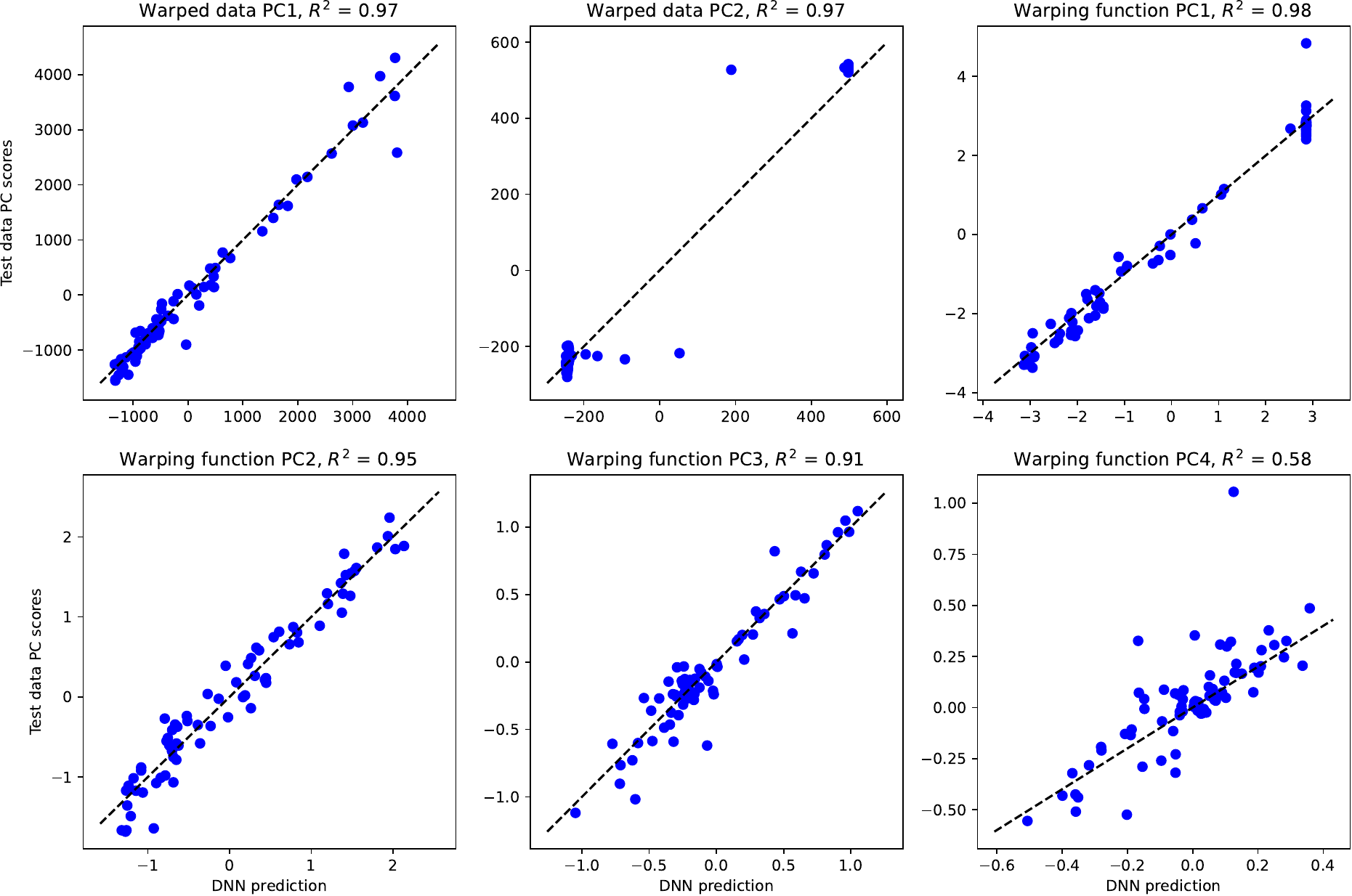}
	\caption[]{Validation of the DNN surrogate model predictions for Method 3.}
	\label{figure:fig_surrogate_Validation_method3}
\end{figure}

Figure \ref{figure:fig_surrogate_Validation_method1}, \ref{figure:fig_surrogate_Validation_method2}, \ref{figure:fig_surrogate_Validation_method3} show the validation results for methods 1-3, using the testing dataset (note that the validation dataset has been used for hyperparameter tuning). Most of the surrogate models show a good prediction accuracy, with a $R^2$ (the predictivity coefficient) value larger than 0.95. However, the GP/DNN model for the fourth PC of conventional PCA (Figures \ref{figure:fig_surrogate_Validation_method1} and \ref{figure:fig_surrogate_Validation_method2}) and the DNN model for the fourth PC of the warping function (Figure \ref{figure:fig_surrogate_Validation_method3}) do not perform as well. Nevertheless, these higher order PCs are not as important as the previous ones due to their much smaller contribution to the total variance, as shown in Figure \ref{figure:fig_PCA_variance}.

\begin{figure}[!htb]
	\centering
	\includegraphics[width=0.8\textwidth]{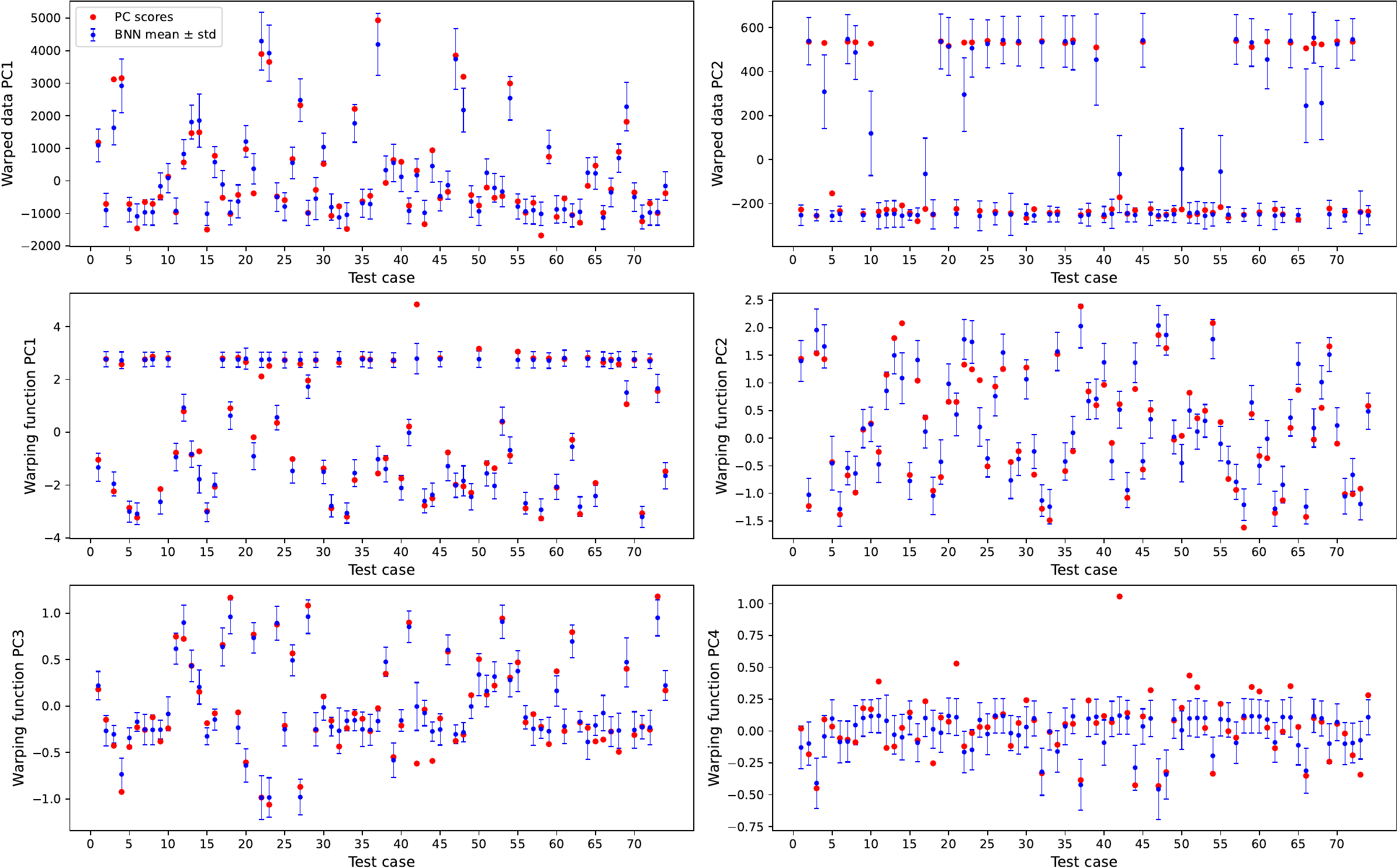}
	\caption[]{Validation of the BNN surrogate model predictions for Method 4.}
	\label{figure:fig_surrogate_Validation_method4}
\end{figure}

Figure \ref{figure:fig_surrogate_Validation_method4} shows the results for BNNs. In the training process using variational inference, the weight parameters are assumed to following Gaussian distributions, whose means and variances are learned. Once a BNN is trained, it can be evaluated multiple times at the same input, each time with different samples of the weight parameters. The resulting predictions can be collected as samples of the responses from which the mean values and variances (uncertainties) can be computed.
To quantify the uncertainty of BNN predictions, we perform 200 predictions for each sample with different network parameters from the posterior distributions of BNN parameters. Figure \ref{figure:fig_surrogate_Validation_method4} presents the mean values and one standard deviations (std in the figure) compared to the test samples. It can be seen that for the majority of the test samples, either the BNN mean values are close to the true PC scores, or the true PC scores are with one standard deviation of the BNN predictions. One exception is the fourth PC of warping function just like Method 3 in Figure \ref{figure:fig_surrogate_Validation_method3}. As discussed above, this is not expected to cause an issue in further analysis.

\begin{figure}[!htb]
	\centering
	\includegraphics[width=0.95\textwidth]{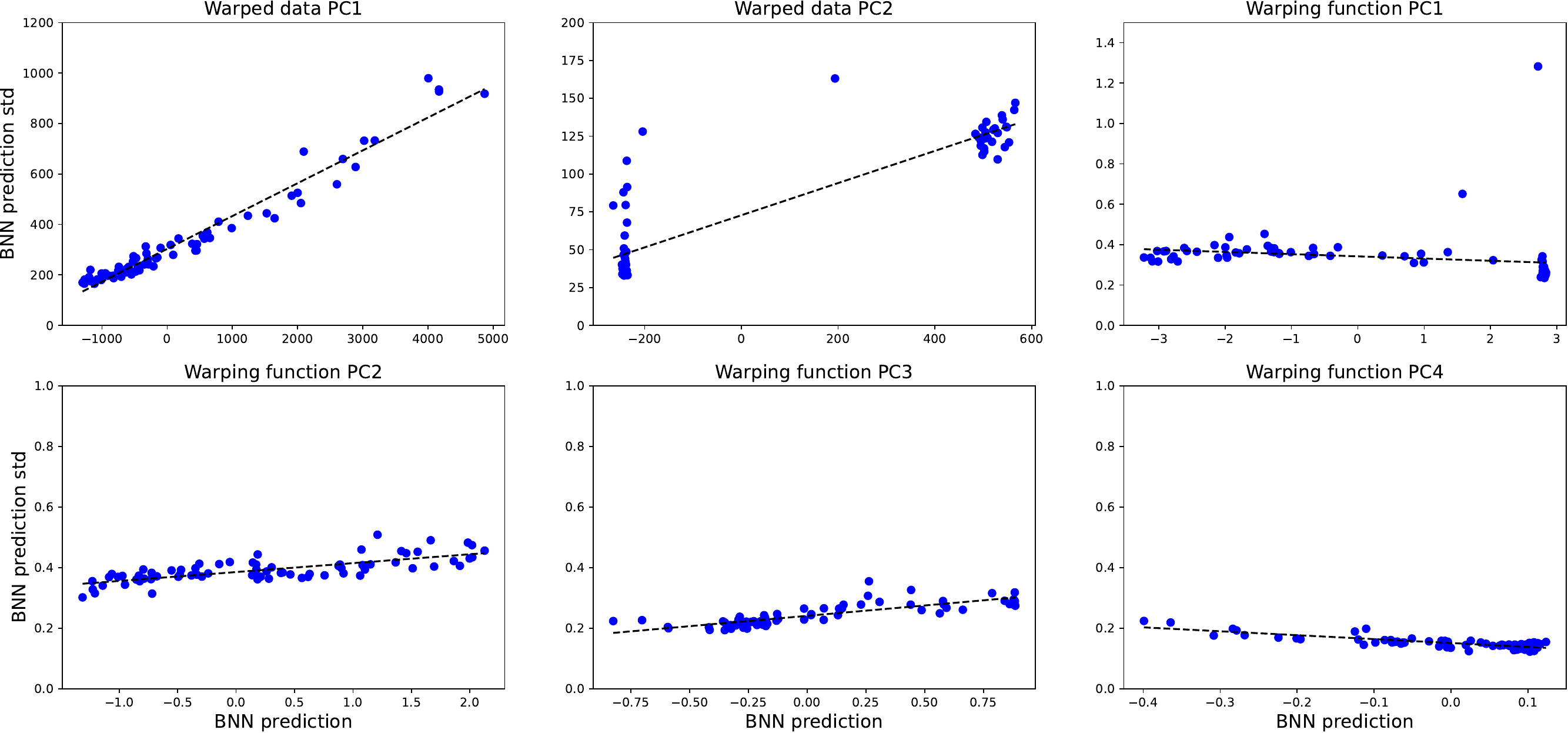}
	\caption[]{Standard deviations of BNN predictions for 6 different PCs.}
	\label{figure:fig_BNN_std_prediction}
\end{figure}

The uncertainties of the BNN predictions will be included in the Bayesian IUQ process as $\bm{\Sigma}_{\text{code}}$. In Method 1, the variance from GP is directly available without further computation. However, this is not true for BNN surrogate models in Method 4, because one needs to run the BNN many times in order to obtain the prediction uncertainties as shown in Figure \ref{figure:fig_surrogate_Validation_method4}. This will significantly slow down the MCMC sampling process and it contradicts our intention to use surrogate models. Figure \ref{figure:fig_BNN_std_prediction} shows the relationship between BNN predictions and the corresponding standard deviations for the testing samples, which indicates linear relationships for most cases. Based on this observation, we have made a simplification by fitting the standard deviations as linear functions of the BNN predictions. During MCMC sampling, the BNNs are only evaluated once for every random walk, and the uncertainties of the surrogate models (in terms of the standard deviations) are evaluated by these linear relations. Note that the test samples for the second PC of the warped data appears to form two clusters instead of a linear relationship. In this case, we simply take the centroids of the two clusters, and determine the standard deviation based on which cluster the BNN prediction is closer to.

\subsection{Results for IUQ}
\label{section:Results3-IUQ}

\begin{figure}
    \centering
    \includegraphics[width=0.95\textwidth]{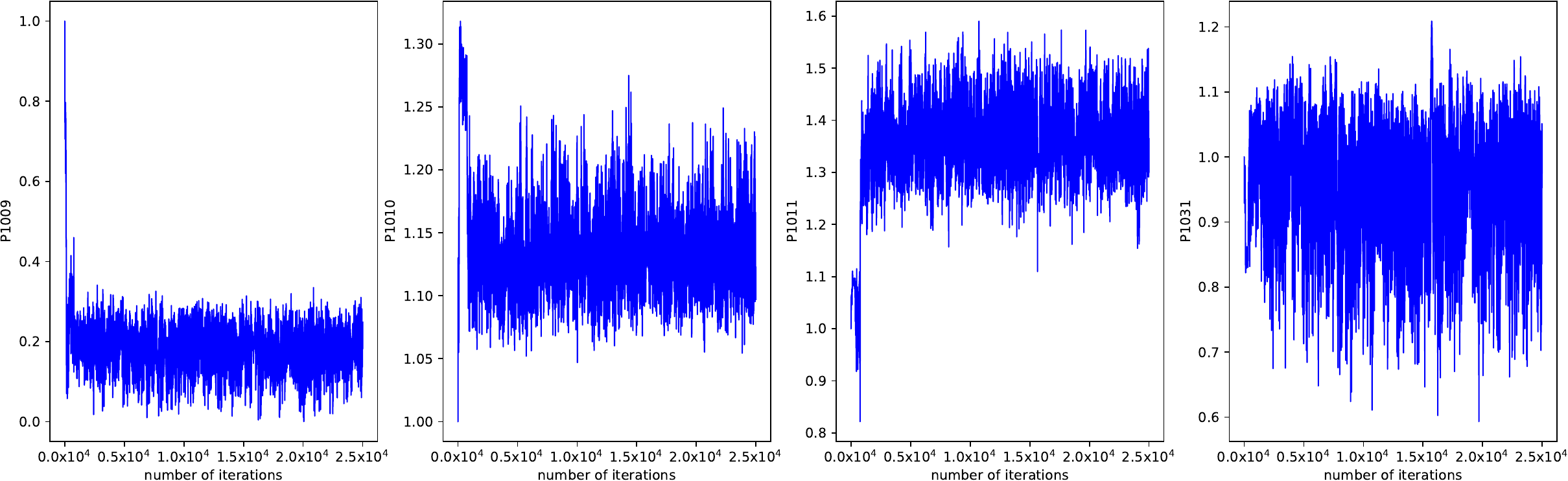}
    \caption{Trace plots of the posterior samples from MCMC sampling for Method 1. The MCMC samples from the other IUQ methods show similar trace plots.}
    \label{figure:fig_IUQ_chain_method3}
\end{figure}

For each IUQ method, 25,000 MCMC samples were generated to explore the posterior distributions of the calibration parameters. The MCMC sampling process takes about 1-2 hours using the surrogate models, which would otherwise take a few thousand hours using the TRACE system code. The first 5,000 samples were abandoned as burn-in since the MCMC chains are not converged in the beginning, which is shown in Figure \ref{figure:fig_IUQ_chain_method3}. Afterwards, we picked every 20 remaining samples for the purpose of ``thinning" to reduce the auto-correlation among the MCMC samples. The remaining 1,000 posterior samples were investigated for the posterior distributions of the calibration parameters.

\begin{figure}
    \centering
    \includegraphics[width=0.7\textwidth]{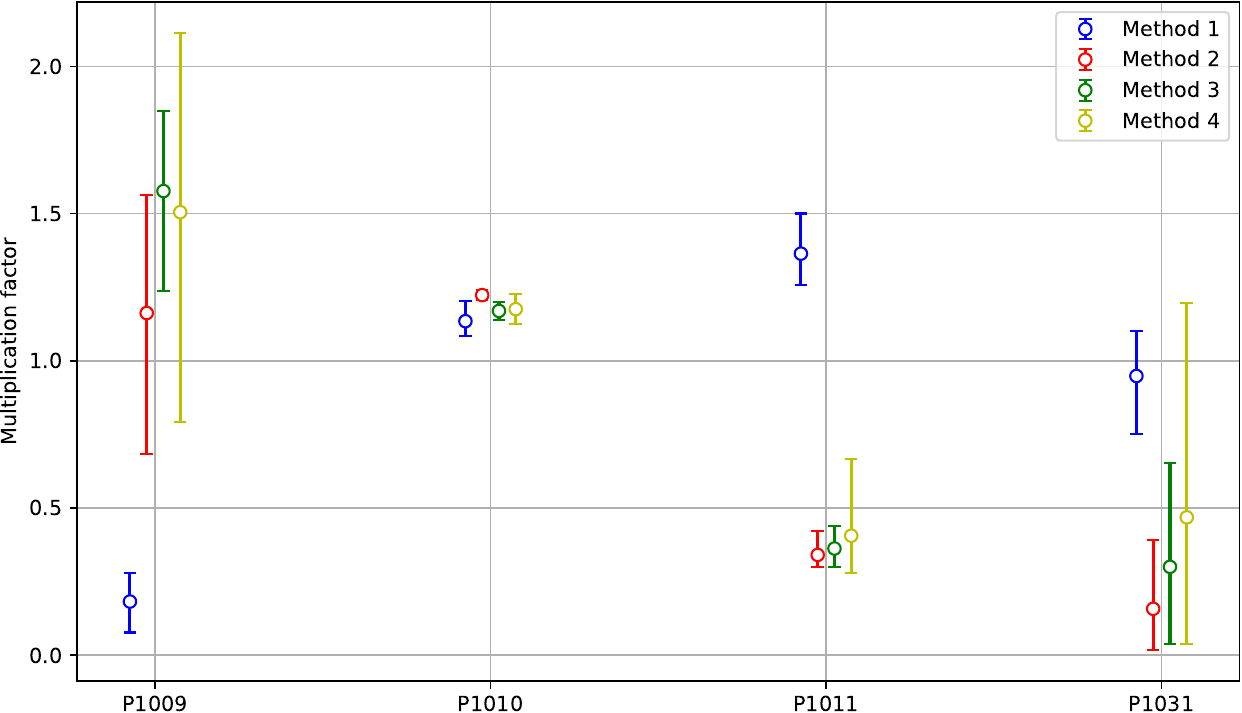}
    \caption{A visual comparison of the mean values and 95\% credible intervals of the calibration parameters with 4 different IUQ methods.}
    \label{figure:fig_IUQ_bar}
\end{figure}

\begin{figure}[!htb]
    \begin{minipage}[b]{.45\linewidth}
        \centering
        \includegraphics[width=0.9\textwidth]{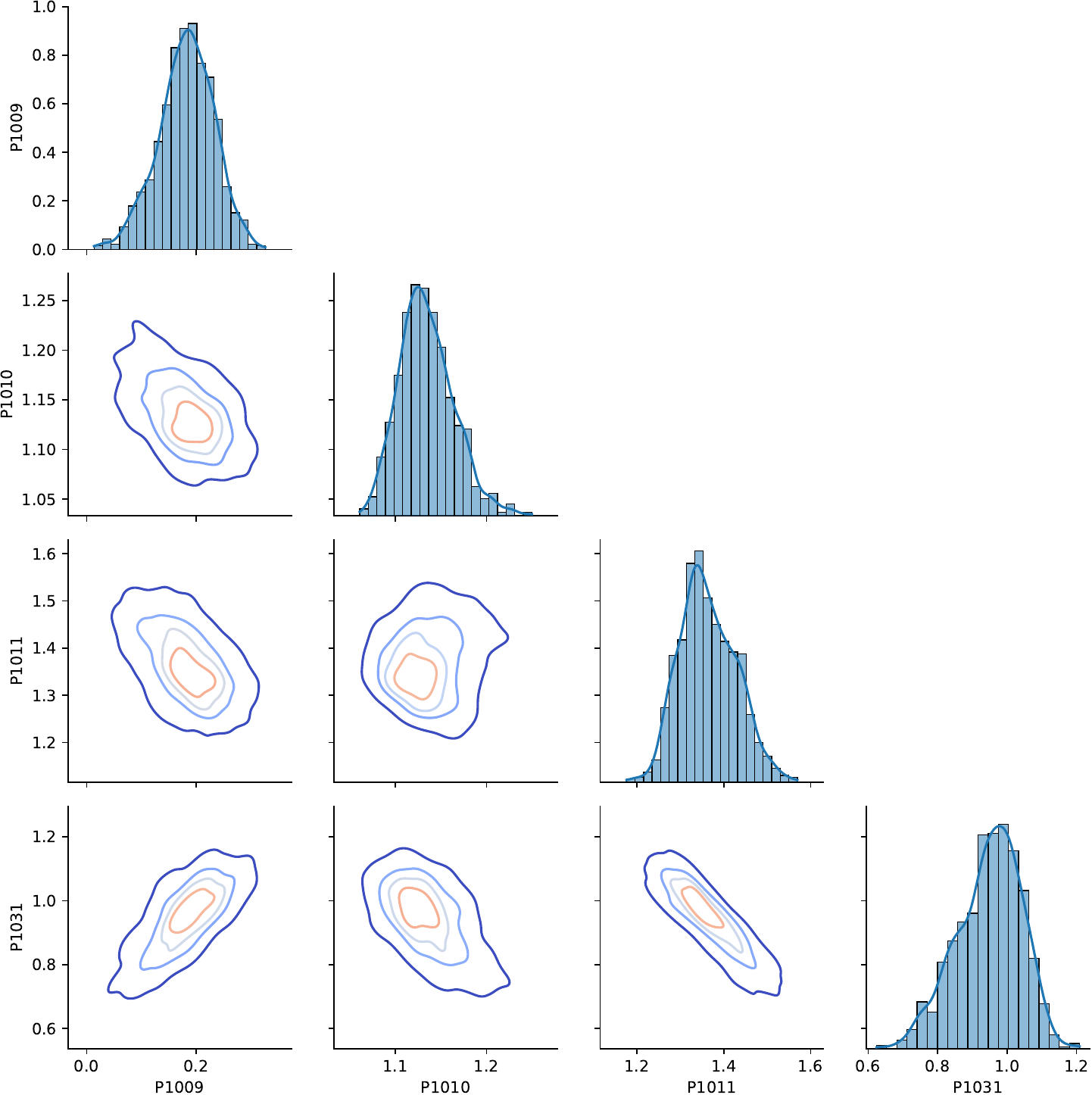}
    \end{minipage}%
    \begin{minipage}[b]{.45\linewidth}
        \centering
        \includegraphics[width=0.9\textwidth]{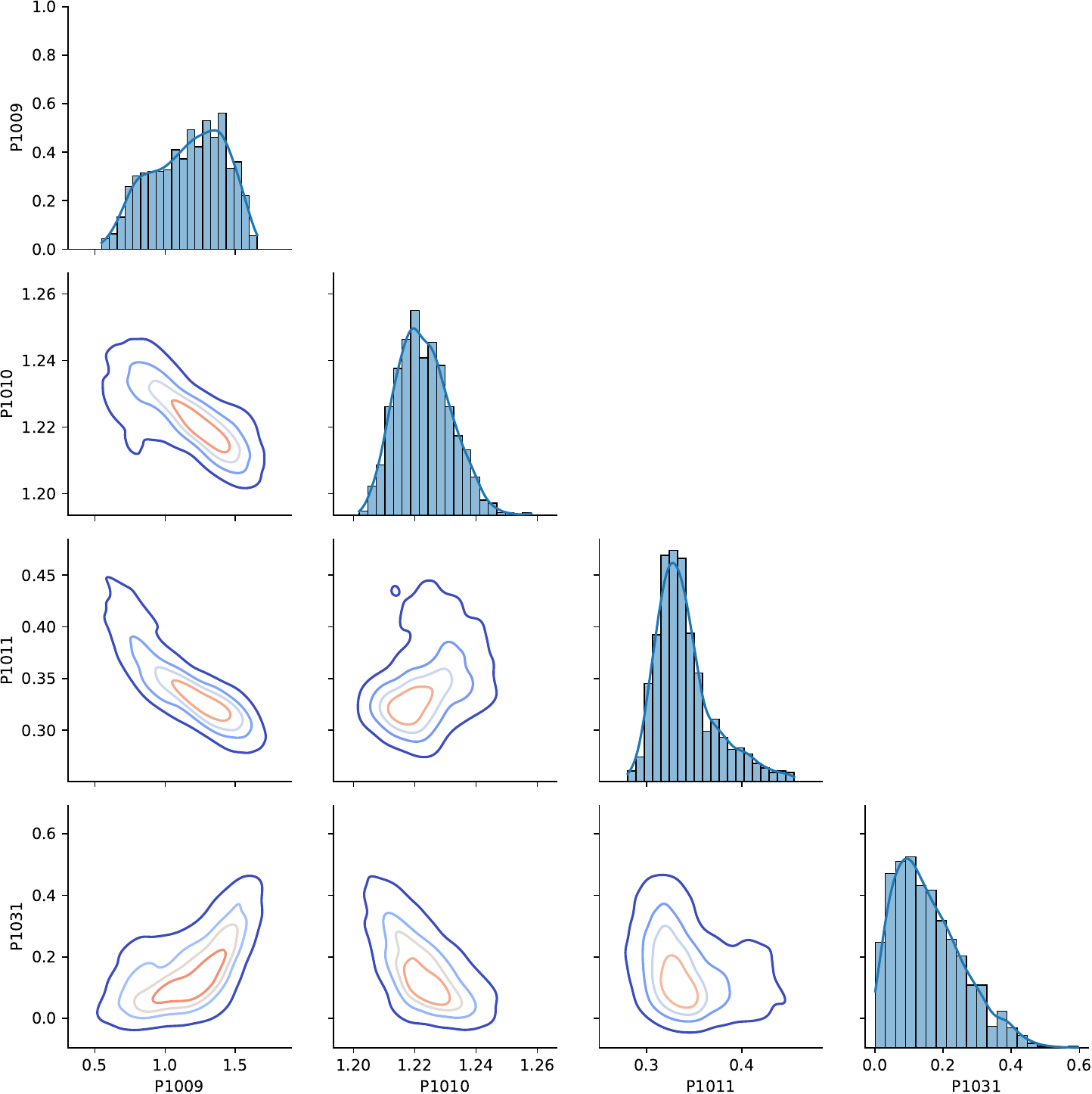}
    \end{minipage}
    \caption{Posterior distributions of the 4 calibration parameters based on MCMC samples, pair-wise joint densities (off-diagonal), and marginal densities (diagonal). Left: Method 1; Right: Method 2.}
    \label{figure:fig_IUQ_contour_method1-2}
\end{figure}

\begin{figure}[!htb]
    \begin{minipage}[b]{.45\linewidth}
        \centering
        \includegraphics[width=0.9\textwidth]{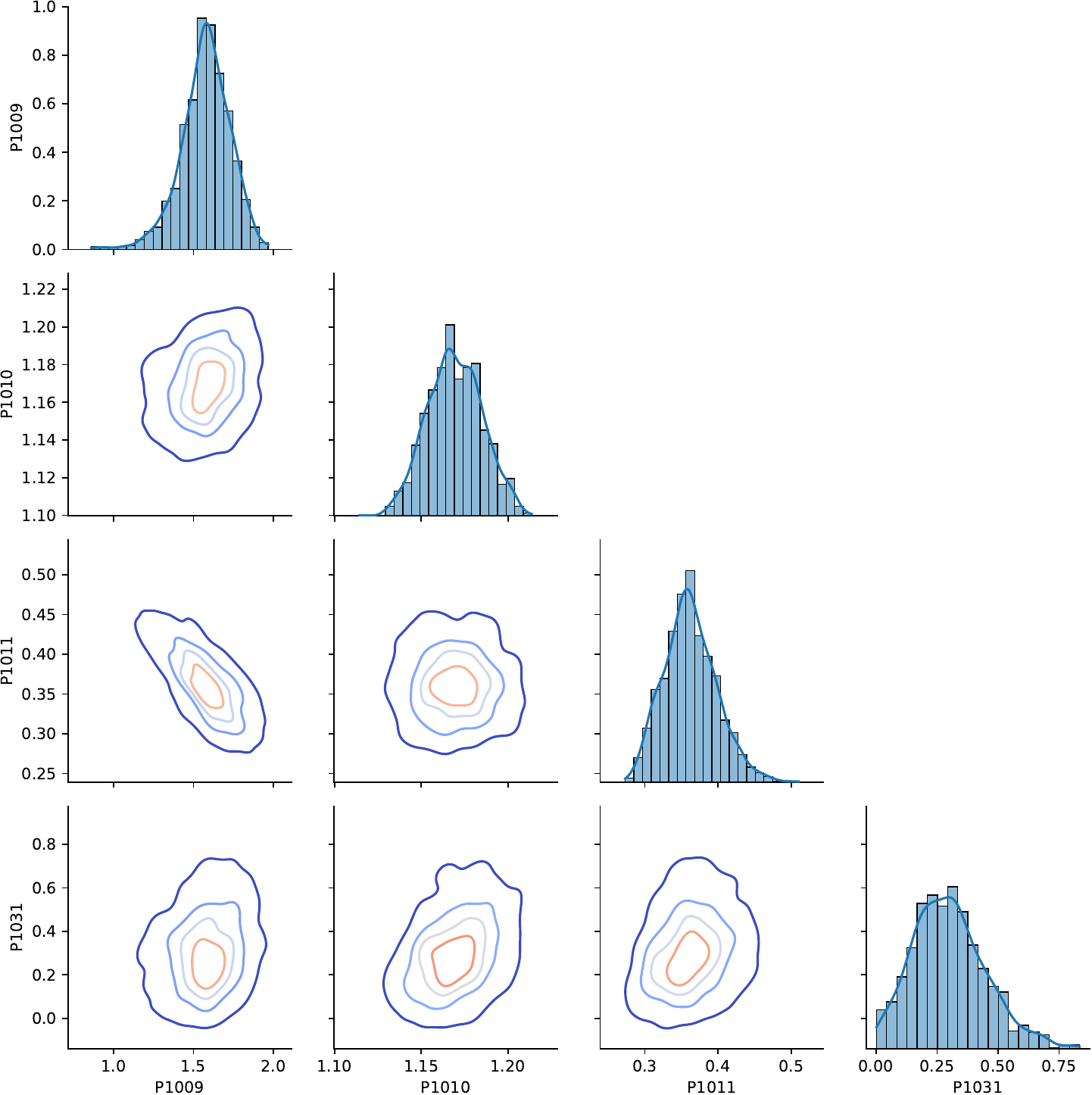}
    \end{minipage}%
    \begin{minipage}[b]{.45\linewidth}
        \centering
        \includegraphics[width=0.9\textwidth]{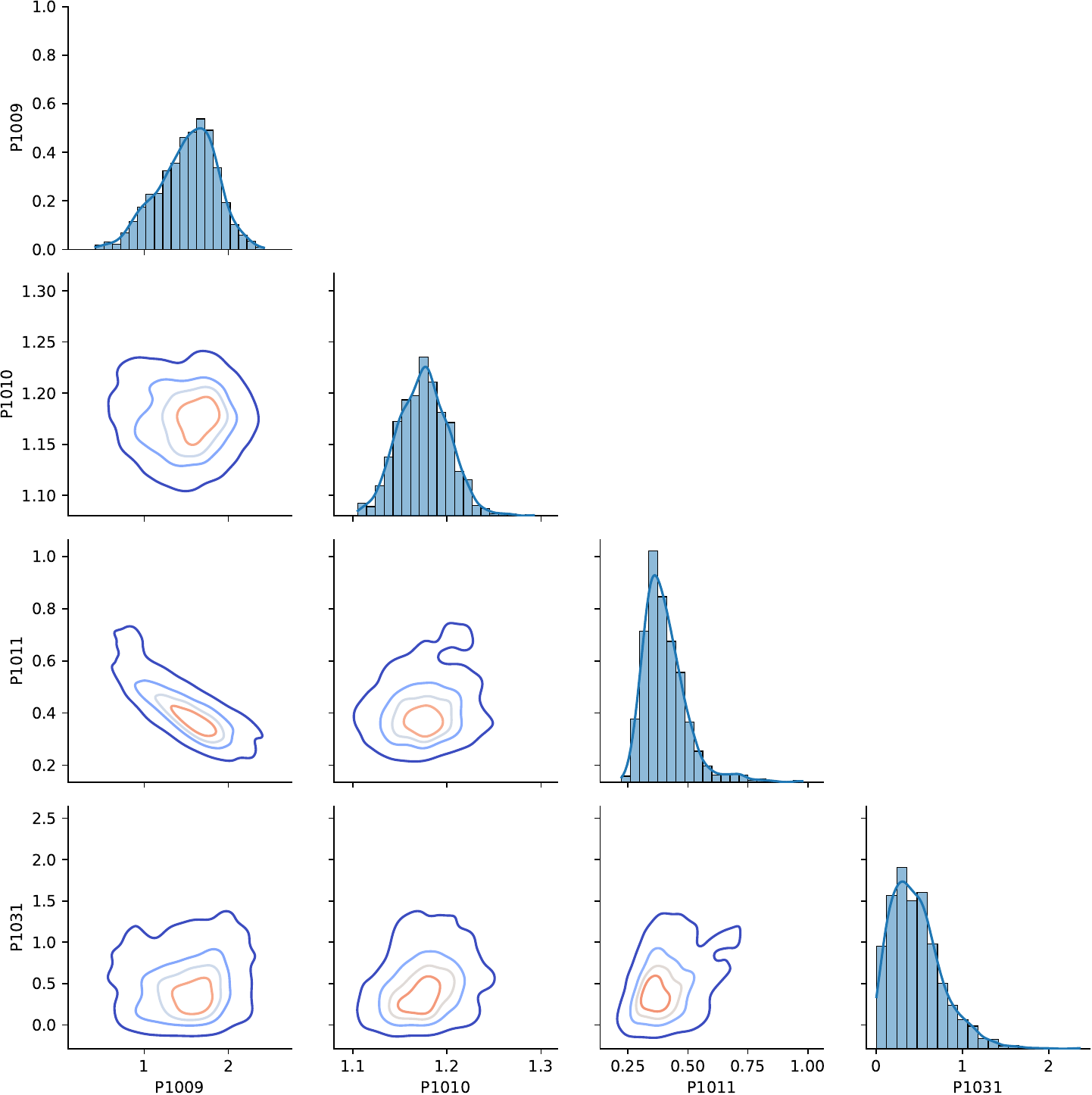}
    \end{minipage}
    \caption{Posterior distributions of the 4 calibration parameters based on MCMC samples, pair-wise joint densities (off-diagonal), and marginal densities (diagonal). Left: Method 3; Right: Method 4.}
    \label{figure:fig_IUQ_contour_method3-4}
\end{figure}

Table \ref{table:IUQ_results} and Figures \ref{figure:fig_IUQ_bar},  \ref{figure:fig_IUQ_contour_method1-2}, and \ref{figure:fig_IUQ_contour_method3-4} present the posterior distributions of the four calibration parameters by all the four IUQ methods. For Method 1, the posterior distributions based on the GP model have a larger differences from other methods. A potential reason for this could be the prediction accuracy for Methods 2-4 based on DNN surrogate models are better. For parameter \texttt{P1010}, which is the most sensitive parameter among the four (see reference \cite{xie2022bayesian-physor} for more details), the mean values between different methods are similar, which gives similar results in the PCT profile (as shown in Section \ref{section:Results4-FUQ-validation}). The posterior results for all four methods have demonstrated a significant reduction of uncertainty from the prior distributions. Compared with Method 2 and 3, the results from Method 4 have a larger uncertainty, since the code uncertainty from the BNN models is considered.

\begin{table}[!htb]
	\centering
	\footnotesize
	\caption{Posterior mean values, standard deviations and 95\% credible intervals for all the calibration parameters using different IUQ methods.}
	\label{table:IUQ_results} 
	\begin{tabular}{l c c c c}
		\toprule
		Methods & Representations & Mean values  & Standard deviations  & 95\% credible intervals\\ 
		\midrule
		\multirow{4}*{Method 1} & \texttt{P1009} & 0.189 & 0.0502 & [0.077, 0.278]\\
		~                      & \texttt{P1010} & 1.136 & 0.0295 & [1.084, 1.201]\\
		~                      & \texttt{P1011} & 1.352 & 0.0638 & [1.256, 1.150]\\
		~                      & \texttt{P1031} & 0.952 & 0.0922 & [0.750, 1.102]\\
		\midrule
		\multirow{4}*{Method 2} & \texttt{P1009} & 1.162 & 0.2532 & [0.684, 1.564]\\
		~                      & \texttt{P1010} & 1.223 & 0.0087 & [1.208, 1.241]\\
		~                      & \texttt{P1011} & 0.340 & 0.0314 & [0.298, 0.422]\\
		~                      & \texttt{P1031} & 0.157 & 0.1015 & [0.016, 0.391]\\
		\midrule
		\multirow{4}*{Method 3} & \texttt{P1009} & 1.576 & 0.1529 & [1.235, 1.849]\\
		~                      & \texttt{P1010} & 1.169 & 0.0156 & [1.139, 1.200]\\
		~                      & \texttt{P1011} & 0.362 & 0.0360 & [0.299, 0.439]\\
		~                      & \texttt{P1031} & 0.300 & 0.1498 & [0.038, 0.652]\\
		\midrule
		\multirow{4}*{Method 4} & \texttt{P1009} & 1.504 & 0.3445 & [0.792, 2.114]\\
		~                      & \texttt{P1010} & 1.175 & 0.0266 & [1.126, 1.226]\\
		~                      & \texttt{P1011} & 0.406 & 0.0938 & [0.280, 0.667]\\
		~                      & \texttt{P1031} & 0.468 & 0.3086 & [0.038, 1.197]\\
		\bottomrule
	\end{tabular}
\end{table}

One advantage of the Bayesian IUQ method is that it can identify the correlations between different calibration parameters through random walk of the MCMC samples, even though the prior distributions are assumed to be independent. The parameters' marginal distributions and pair-wise joint distributions are shown in Figures \ref{figure:fig_IUQ_contour_method1-2} and \ref{figure:fig_IUQ_contour_method3-4}. For examples, for all of the four IUQ methods, \texttt{P1009} and \texttt{P1011} have a strong negative correlation. As a result, when generating new samples from posterior distributions, the correlations between the calibration parameters should be considered. There are some differences in the posterior marginal/joint distributions by different IUQ methods, this is expected since inverse problems are usually ill-posed with many different solutions. To validate the IUQ results, we will determine whether the posterior distributions will make TRACE simulations more consistent with the FEBA experimental data, not only with the test case that has been used in IUQ, but also test cases unseen during IUQ.

\subsection{Results for FUQ and validation}
\label{section:Results4-FUQ-validation}

To determine which IUQ method produces the best IUQ results, we propagated the quantified posterior distributions of the calibration parameters through the TRACE model to obtain the updated prediction uncertainties in the PCT profiles. This step can take advantage of the existing GP/DNN surrogate models that were trained during the IUQ process to reduce the computational cost in the FUQ process. We generated 1000 random samples from the joint distributions of the posterior distributions from each IUQ method, then we used the surrogate models in each IUQ method to generate the PC scores and subsequently the reconstructed PCT profiles. 

\begin{figure}[!htb]
	\centering
	\includegraphics[width=0.99\textwidth]{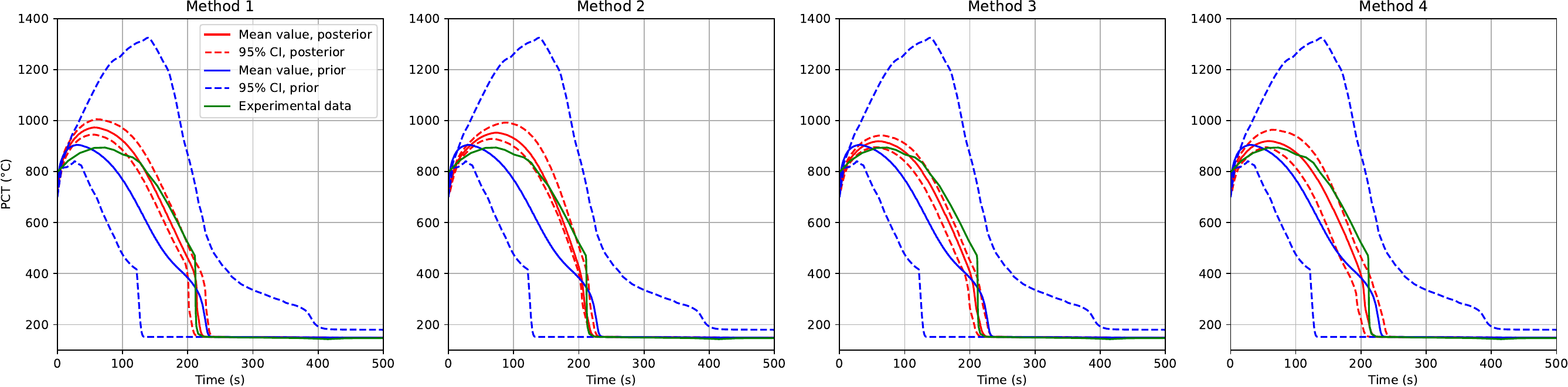}
	\caption[]{TRACE simulations based on posterior and prior distributions and their 95\% confidence intervals, compared with the FEBA experiment data at $z = 2225$ mm. This is the test case whose data has been used for IUQ.}
	\label{figure:fig_FUQ_2225}
\end{figure}

The results of the FUQ process based on both the posterior and prior distributions at axial position $z = 2225$ mm are shown in Figure \ref{figure:fig_FUQ_2225}, together with the FEBA experimental data. Note that this is only a proof-of-concept of Bayesian IUQ, rather than a valid ``validation'' process, because the same data has already been used in IUQ. It can be observed that: (1) the 95\% confidence intervals based on the posteriors are much smaller than those based on the prior, due to reduction of uncertainty by IUQ, (2) compared with Methods 1/2 that used conventional PCA, the FUQ results of Methods 3/4 with fPCA have a better agreement with the experimental data, especially around $t_{\text{max}}$. After considering the code uncertainty using BNN surrogate models, the FUQ results of Method 4 show a larger 95\% confidence interval than Method 3, which covers most of the experimental data.

\begin{figure}[!htb]
	\centering
	\includegraphics[width=0.99\textwidth]{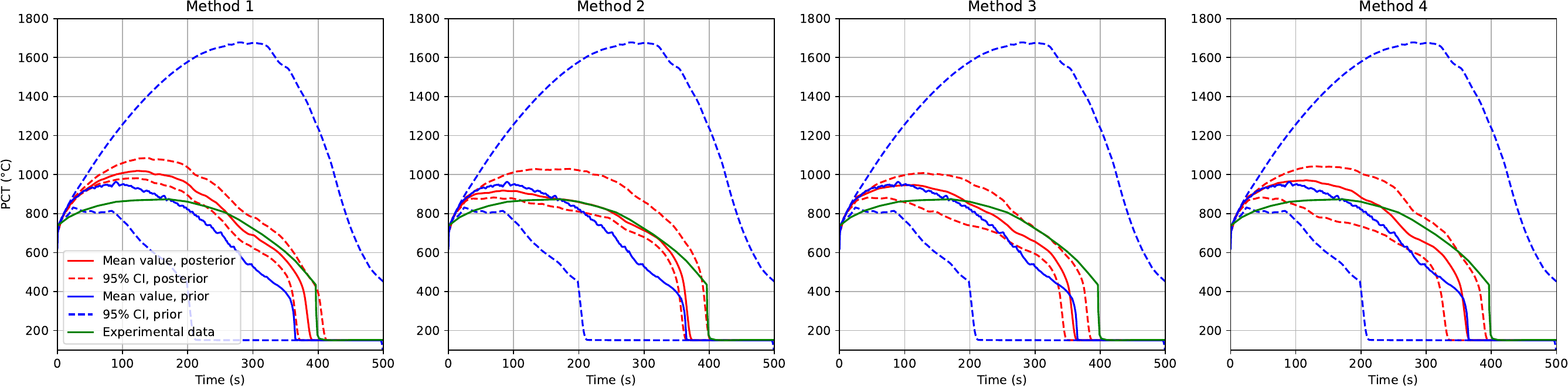}
	\caption[]{TRACE simulations based on posterior and prior distributions and their 95\% confidence intervals, compared with the FEBA experiment data at $z = 1135$ mm. This is a test case unseen during IUQ.}
	\label{figure:fig_FUQ_1135}
\end{figure}

\begin{figure}[!htb]
	\centering
	\includegraphics[width=0.99\textwidth]{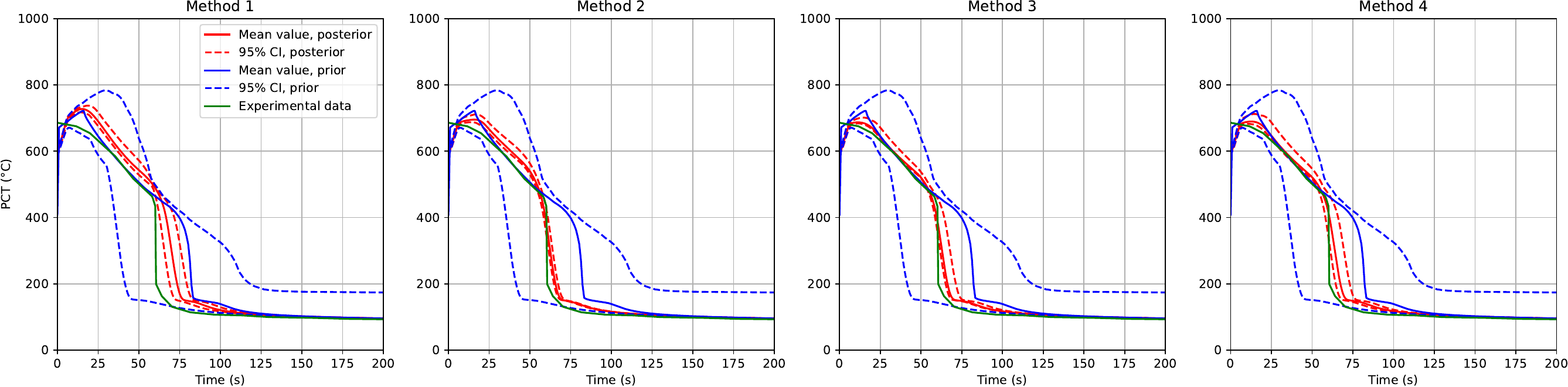}
	\caption[]{TRACE simulations based on posterior and prior distributions and their 95\% confidence intervals, compared with the FEBA experiment data at $z = 3315$ mm. This is a test case unseen during IUQ.}
	\label{figure:fig_FUQ_3315}
\end{figure}

To perform a more rigorous validation of the IUQ results, new experimental data that is not seen during IUQ should be used. We performed FUQ and validation at two other axial positions ($z = 1135$ mm and $3315$ mm) of the FEBA experiment test 216. Since the surrogate models are not applicable for these datasets, we need to use TRACE to run the samples used for FUQ. We generated 300 random samples from joint posterior distributions of the four different IUQ methods and ran the TRACE model to get the PCT profiles. The FUQ and validation results at the two axial positions are shown in Figure \ref{figure:fig_FUQ_1135} and \ref{figure:fig_FUQ_3315}, respectively. For all of the FUQ results, the mean values based on the posteriors show a better agreement with the experimental data. Other observations are similar with Figure \ref{figure:fig_FUQ_2225}. Methods 1/2 produce results that have larger disagreement with the data before and around $t_{\text{max}}$, while Methods 3/4 produce results that have slightly larger disagreement with the data at around $t_{\text{quench}}$. Overall, the FUQ results from Method 4 has the largest posterior uncertainty range and the best coverage of the experimental data. This proves that the combination of fPCA and DNN-based surrogate model while accounting for the code uncertainty has improved the Bayesian IUQ process for this transient dataset.

\section{Conclusions}
\label{section:Conclusions}

This paper proposed a Bayesian inverse Uncertainty Quantification (IUQ) process for time-dependent responses using four methods that used different dimensionality reduction processes and surrogate models. We proposed a framework for Bayesian IUQ that combine functional principal component analysis (PCA) and deep neural network (DNN)-based surrogate models while accounting for the code/interpolation uncertainty. Functional PCA separates the phase and amplitude information of the time series data before dimensionality reduction, which shows an improved performance over the conventional method. The use of DNN-based surrogate models has also proven to be very effective in representing the PC scores, and it significantly reduces the computational cost in Markov Chain Monte Carlo (MCMC) sampling sampling. We also considered code uncertainty for surrogate models in Bayesian IUQ by adopting Bayesian neural networks (BNNs). Since the sampling-based UQ process for BNN will increase the computation cost in the IUQ process, we estimate the BNN uncertainty with a linear regression model since there is a clear linear relationship between BNN prediction and uncertainty. The proposed approach has been applied to the peak cladding temperature in the FEBA benchmark. Forward Uncertainty Quantification (FUQ) and validation of the proposed IUQ method have demonstrated that the code simulations based on the posterior distributions have an improved agreement with the experimental data while the uncertainty ranges can envelop the majority of the experimental data.

The primary limitation of this framework is that the model uncertainty is not considered in this IUQ study since there is only one experimental test is considered. In further study, we will include the model discrepancy term that comes from the missing and inaccurate physics in the system code, in order to design a more comprehensive IUQ process. We seek to find a mathematical representation for FEBA transient data. In addition, an IUQ method based on hierarchical Bayesian modeling can be applied since data at different axial position can be considered through this model.


\bibliography{./bibliography.bib}

\end{document}